\documentclass[runningheads]{llncs}

\usepackage[year=2026]{eccv}

\usepackage{eccvabbrv}

\usepackage{graphicx}
\usepackage{float}
\usepackage{enumitem}
\usepackage{booktabs}
\usepackage{kotex}
\usepackage{colortbl}      
\usepackage{xcolor}        
\usepackage{multirow}      
\usepackage{array}       
\usepackage{amssymb}
\usepackage[table]{xcolor}
\usepackage{pifont}
\usepackage{booktabs}
\usepackage{makecell}
\usepackage{subcaption}

\usepackage{setspace}
\usepackage{tcolorbox}
\tcbuselibrary{skins, breakable}

\usepackage{algorithm}
\usepackage{algpseudocode}

\usepackage[misc]{ifsym}

\usepackage{algorithm}
\usepackage{algpseudocode}
\usepackage[accsupp]{axessibility}  

\usepackage{hyperref}

\usepackage{orcidlink}

\begin{document}

\title{3DZip: Spatial-Aware Feature Diversity-Guided Token Compression for 3D Question Answering} 

\titlerunning{3DZip}

\author{Changwoo Baek\orcidlink{0009-0008-2616-8238} \and
Kyeongbo Kong\textsuperscript{\Letter}\orcidlink{0000-0002-1135-7502}}

\authorrunning{Baek and Kong}


\institute{Department of Electrical and Electronics Engineering, Pusan National University,
Busan, Republic of Korea
\\
\email{\{higok18, kbkong\}@pusan.ac.kr}\\
\vspace{2pt}
\url{https://cvsp-lab.github.io/3DZip}
}

\maketitle

\begingroup
\renewcommand{\thefootnote}{}
\footnotetext{\textsuperscript{\Letter} Corresponding author.}
\endgroup

\begin{abstract}
Recent 3D vision-language models (3D VLMs) construct geometry aware tokens by projecting 2D visual features into world coordinates, enabling spatial reasoning for tasks such as 3D question answering. However, this design generates thousands of tokens per scene, resulting in substantial computational and memory overhead. While token compression has been extensively studied in 2D VLMs, existing approaches rely on semantic relevance or attention-based selection that overlook the structured spatial nature of 3D tokens. Moreover, redundancy in 3D representations cannot be resolved by spatial proximity alone, as object-level token imbalance persists even after spatial aggregation. To address this, we propose 3DZip, a three-stage token compression framework that first applies coarse voxelization to remove point-level redundancy, then selects anchor tokens based on feature-space diversity via a Determinantal Point Process, and finally merges remaining tokens under spatial constraints to preserve geometric coherence. Experiments on three 3D question answering benchmarks demonstrate that 3DZip consistently outperforms existing compression methods, retaining 94.7\% of the original performance with only 128 tokens, achieving a $1.92\times$ faster inference speed.

\end{abstract}

\section{Introduction}
\label{sec:intro}
Recent advances in large language models (LLMs)~\cite{liu2024deepseek,touvron2023llama} and vision-language models (VLMs)~\cite{bai2023qwen,wang2024qwen2,liu2024llava} have extended multimodal understanding beyond 2D image perception to 3D Vision-Language Models (3D VLMs)~\cite{zhu2024llava,deng20253d,huang2024chat,leo}, which are capable of reasoning over complex three-dimensional environments. In particular, 3D Question Answering~\cite{ma2022sqa3d,majumdar2024openeqa,azuma2022scanqa}, which requires understanding object attributes, spatial relations, and scene-level context from natural language queries, has emerged as a key task for embodied AI and 3D robotics applications.

Early attempts to address 3D Question Answering relied on point cloud representations~\cite{deng20253d}. However, the lack of large-scale 3D language datasets and powerful pretrained 3D encoders limited their performance compared to approaches leveraging strong 2D visual backbones. As a result, recent works adopt a projection-based paradigm that lifts 2D visual features into a shared 3D coordinate system using depth and camera pose information~\cite{zhu2024llava}. This approach allows models to reuse powerful 2D pretrained representations while incorporating spatial structure for 3D reasoning.

\begin{figure}[t]
    \centering
    \includegraphics[width=1\linewidth]{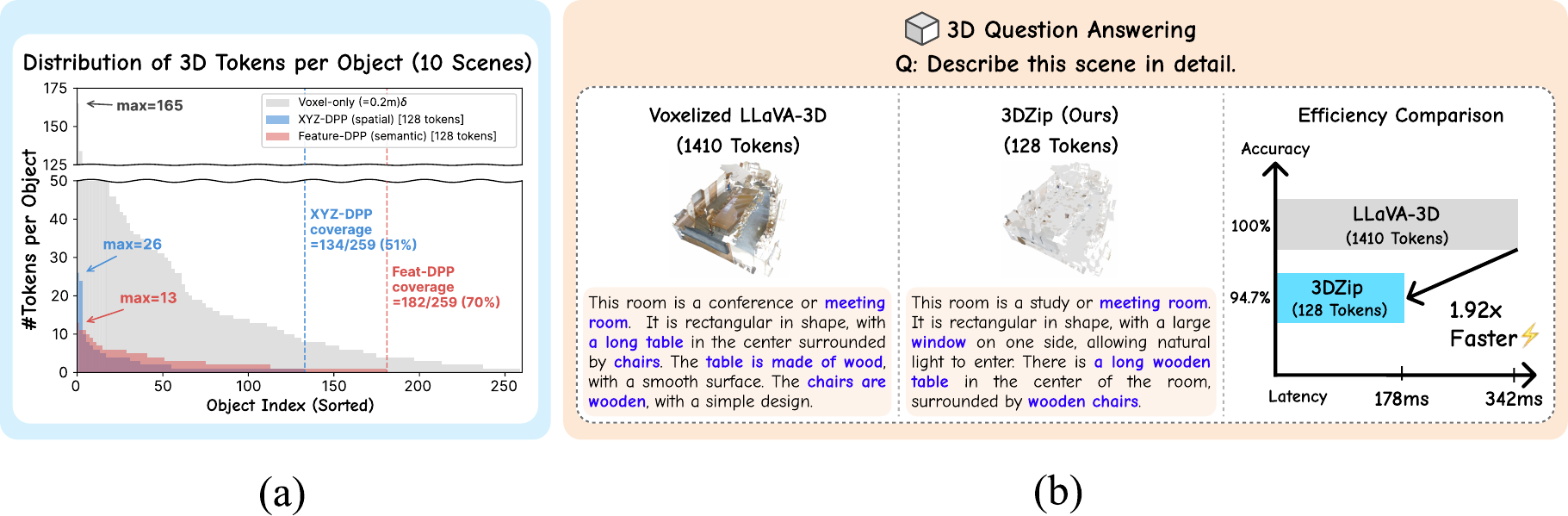}
    \vspace{-17pt}
   \caption{
    \textbf{(a) Per-object token allocation under different selection strategies.} Object coverage is defined as the fraction of GT object instances that contain at least one selected token, computed using GT object masks and aggregated over 10 scenes from SQA3D. Spatial sampling (XYZ-DPP) concentrates tokens on large objects (51\% coverage), whereas Feature-DPP distributes tokens more evenly across objects, increasing coverage to 70\%.
    \textbf{(b) 3DZip overview.} 3DZip compresses dense 3D tokens to only 128 while retaining 94.7\% accuracy and achieving 1.92$\times$ faster inference.}
    \label{fig:teasor}
    \vspace{-17pt}
\end{figure}

Despite these advantages, projection-based multi-view aggregation introduces two distinct forms of redundancy in 3D VLM representations. First, \textit{point-level redundancy} arises when identical physical surfaces are repeatedly observed across multiple viewpoints. When lifted into a shared 3D coordinate system, these repeated observations generate overlapping tokens that correspond to nearly identical spatial locations. This type of redundancy can be partially mitigated through voxel-based aggregation, which collapses tokens occupying similar 3D positions. Second, \textit{object-level redundancy} occurs when a single object is represented by tokens corresponding to multiple surface regions. For example, an object such as a sofa may produce tokens from its front, side, and back surfaces when observed from different viewpoints. Although these tokens describe the same physical instance, their spatial coordinates remain separated due to the object's geometric extent, making it difficult to collapse them into a single representation using spatial aggregation alone.

As a consequence, object-level redundancy often leads to a highly imbalanced token distribution across objects. As illustrated in Fig.~\ref{fig:teasor}a, even after voxel aggregation, token allocation across objects exhibits a pronounced long-tailed distribution in which a small subset of objects dominates the token budget. Spatially driven diversity strategies such as XYZ-DPP alleviate this issue only partially, as tokens remain concentrated on a limited subset of objects, covering only 51\% of object instances in our analysis. In contrast, selecting tokens based on feature-level diversity produces a markedly more balanced distribution, increasing object coverage to 70\%. These observations suggest that spatial aggregation alone is insufficient to mitigate object-level redundancy in multi-view 3D VLM representations.

Motivated by these observations, we design a structured token compression framework that explicitly addresses different sources of redundancy in multi-view 3D representations. Specifically, we decompose redundancy into three complementary components: point-level density, object-level duplication, and geometric consistency. First, we apply voxel-based spatial aggregation to reduce point-level redundancy caused by multi-view projection. By merging tokens within fixed-size voxels, this step removes tokens corresponding to nearly identical 3D locations while preserving the coarse spatial structure of the scene. Second, we perform feature-level diversity selection to suppress object-level duplication. As discussed above, spatial grouping alone cannot reliably determine whether tokens belong to the same object instance. Instead, selecting tokens based on feature diversity enables the model to retain semantically distinct objects while avoiding repeated representations of the same object surfaces. Finally, the remaining tokens are merged to their nearest anchors according to spatial proximity. This spatially constrained merging step preserves geometric consistency between tokens and maintains the structural relationships required for spatial reasoning.

Based on this design, we propose \textbf{3DZip} (Fig.~\ref{fig:teasor}b), a geometry-aware token compression framework for multi-view 3D VLMs. By aligning token selection with the redundancy structure induced by multi-view aggregation, our method significantly improves efficiency while maintaining strong reasoning performance. Extensive experiments on three 3D question answering benchmarks demonstrate that 3DZip consistently outperforms existing token compression methods. Our approach retains 94.7\% of the original performance using only 128 tokens, achieving a $1.92\times$ faster inference speed, highlighting its effectiveness for scalable multi-view 3D VLM deployment.

\noindent In summary, our contributions are three-fold:
\begin{itemize}
\item We empirically reveal a severe object-level token imbalance induced by projection based multi-view aggregation in 3D VLMs, which persists even after spatial aggregation.
\item We show that spatial grouping alone cannot resolve this imbalance and demonstrate that feature-level diversity is crucial for balancing object coverage in multi-view 3D scenes.
\item We propose \textbf{3DZip}, a structured three-stage token compression framework combining voxel aggregation, feature-diversity–based anchor selection, and spatially constrained merging.
\end{itemize}

\section{Related Works}

\subsection{3D Vision-Language Models}
Following the success of 2D VLMs~\cite{bai2023qwen,liu2024llava}, research has shifted toward extending LLMs to 3D scene understanding. Existing approaches vary by their representation strategies. Prior to LLM-based methods, transformer-based 3D vision-language models such as 3D-VisTA~\cite{zhu20233d} align 3D scenes and text through pre-training and are competitive on 3D question answering. Building on LLMs, early works such as LL3DA~\cite{chen2024ll3da}, LEO~\cite{leo}, and Chat-Scene~\cite{huang2024chat} directly integrate 3D point clouds into LLMs~\cite{deng20253d,man2024situational,qi2024gpt4point,shi2024aware,wang2023c3d,xu2024pointllm}, and Robin3D~\cite{kang2025robin3d} further improves such 3D LLMs through robust instruction tuning. While effective for geometric modeling, these methods rely on specialized 3D encoders and large-scale 3D vision-language datasets and benchmarks~\cite{ma2022sqa3d,azuma2022scanqa,majumdar2024openeqa,lyu2024mmscan}, limiting their scalability and generalization.

To address this, recent studies have adopted projection-based paradigms that leverage pretrained 2D models~\cite{3dllm,scenellm}. Early efforts like 3D-LLM~\cite{3dllm} and Scene-LLM~\cite{scenellm} aggregate 2D features into 3D space but depend on complex pipelines and external segmentation, which are computationally intensive and cause information loss.

In contrast, LLaVA-3D~\cite{zhu2024llava} offers a more unified solution by integrating 3D positional embeddings into 2D patches. This design enables explicit 3D modeling without dedicated encoders, preserving the original LMM's reasoning capabilities. However, LLaVA-3D generates thousands of tokens per scene, leading to substantial computational and memory overhead. Thus, efficient token compression is essential to make these powerful 3D VLMs practical for large-scale deployment.

\vspace{-10pt}
\subsection{Token Compression}
Token compression has been widely studied in VLMs to reduce attention cost and memory overhead caused by a large number of visual tokens. Most existing methods are developed for 2D image–text settings, where visual tokens are assumed to be highly redundant~\cite{fastv,zhang2024sparsevlm,vispruner,yang2025visionzip,FitPrune,lin2025vtw,baekagilepruner,alvar2025divprune,song2026uncertainty,kong2025token}. Early approaches leveraged cross-attention distributions within the LLM, retaining visual tokens that receive high attention from text tokens~\cite{fastv,zhang2024sparsevlm,xing2024pyramiddrop}. However, such methods depend on the decoding process, making them inefficient and difficult to combine with acceleration techniques such as FlashAttention~\cite{dao2022flashattention}. Moreover, attention tends to be biased toward later visual tokens, leading to unstable performance and limited representativeness. To address these issues, recent methods rely on CLS-token attention from the vision encoder~\cite{FasterVLM,vispruner,yang2025visionzip,shang2025prumerge}. While CLS-based selection avoids additional decoding cost and captures globally salient tokens, it often overlooks locally important objects and fine-grained geometric structures. Moreover, these methods are primarily designed for 2D image–text settings and do not explicitly account for 3D spatial structure. To the best of our knowledge, their effectiveness in projection-based 3D VLMs has not been systematically studied.

In projection-based 3D VLMs, multi-view features are lifted into world coordinates to form 3D tokens, substantially increasing token count. While DTC~\cite{dtc} incorporates depth information for token pruning, its focus remains on multi-view 2D VLM settings that do not explicitly learn 3D spatial representations. In contrast, our method jointly considers feature-space diversity and geometric constraints, preserving both semantic representativeness and spatial coherence in projection-based 3D VLMs.

\begin{figure*}[t]
    \centering
    \includegraphics[width=1\linewidth]{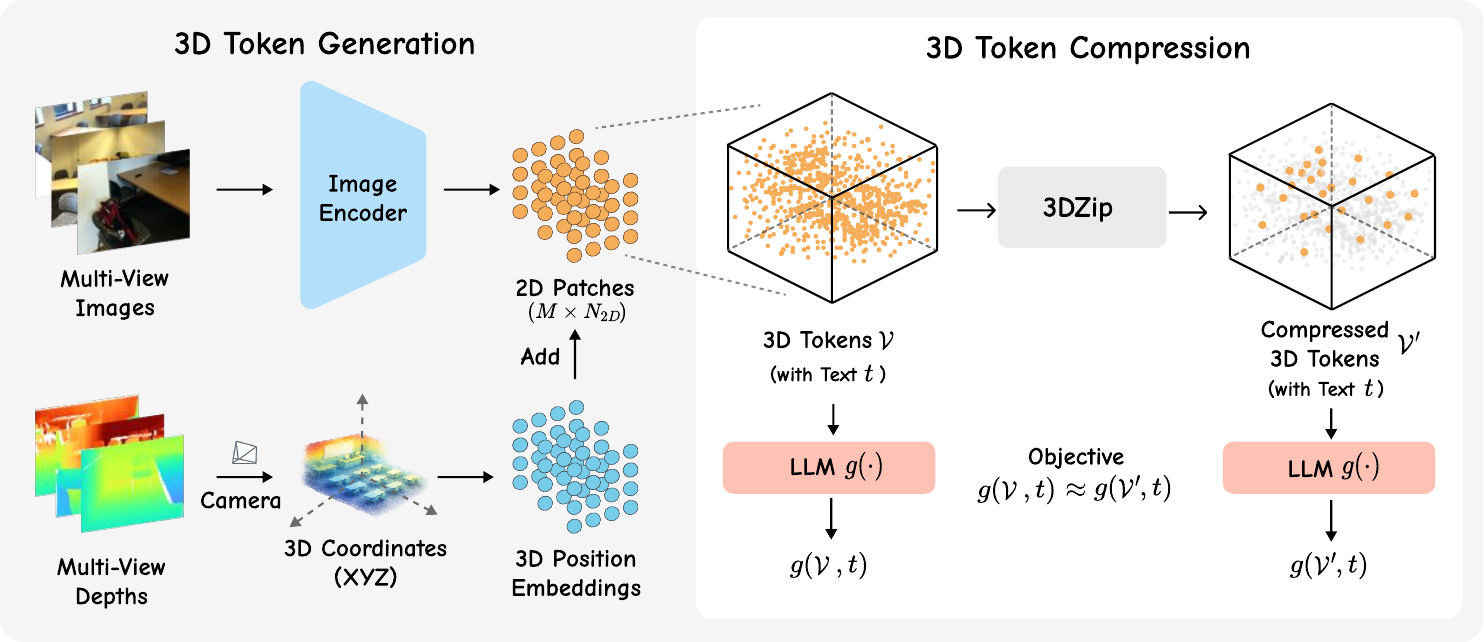}
    \caption{
   \textbf{Overview of geometry-aware 3D token construction and the compression objective.}
    Multi-view RGB-D inputs are projected into world coordinates to form geometry-aware 3D tokens $\mathcal{V}$ via 3D positional embedding.  Due to the large token cardinality $N = M \times N_{2D}$, an efficient token compression strategy is crucial.
    }
    \label{fig:3DVLM}
    \vspace{-10pt}
\end{figure*}

\vspace{-10pt}
\section{Preliminary: Projection-based 3D VLMs}

Projection-based 3D VLMs construct geometry-aware tokens by lifting multi-view 2D visual features into a shared 3D coordinate system. As illustrated in Fig.~\ref{fig:3DVLM}, multi-view RGB-D observations are first processed by a pretrained image encoder to extract 2D visual tokens. Using the corresponding depth maps and camera poses, these tokens are then back-projected into the world coordinate system and augmented with 3D positional embeddings to form geometry-aware 3D tokens. Let $M$ denote the number of views, and each view produces $N_{2D}$ visual tokens extracted by the image encoder. For view $m$, the 2D token features are
\begin{equation}
\{\mathbf{f}_{m,i}\}_{i=1}^{N_{2D}}, 
\quad \mathbf{f}_{m,i} \in \mathbb{R}^d .
\end{equation}

Given the depth map $D_m$ and camera pose $T_m \in SE(3)$, each token is back-projected into the world coordinate system:
\begin{equation}
\mathbf{p}_{m,i} = \Pi^{-1}(\mathbf{u}_{m,i}, D_m, T_m),
\end{equation}
where $\mathbf{u}_{m,i}$ denotes the 2D pixel location corresponding to token $i$. To incorporate geometric information, the 3D coordinate $\mathbf{p}_{m,i}$ is encoded using a learnable positional embedding $\phi(\mathbf{p}_{m,i}) \in \mathbb{R}^d$ and combined with the visual feature:
\begin{equation}
\tilde{\mathbf{f}}_{m,i} = 
\mathbf{f}_{m,i} + \phi(\mathbf{p}_{m,i}).
\end{equation}

Each geometry-aware 3D token is therefore represented as
\begin{equation}
v_{m,i} = (\tilde{\mathbf{f}}_{m,i}, \mathbf{p}_{m,i}).
\end{equation}

Aggregating tokens across all views yields a token set
\begin{equation}
N = M \times N_{2D}.
\end{equation}

For notational simplicity, we flatten the token indices and denote the resulting 3D tokens as
\begin{equation}
v_i = (\tilde{\mathbf{f}}_i, \mathbf{p}_i), \quad i = 1,\dots,N .
\end{equation}

Let the original token set be
\begin{equation}
\mathcal{V} = \{v_i\}_{i=1}^{N}.
\end{equation}

Our goal is to construct a compressed token set

\begin{equation}
 \quad |\mathcal{V}'| = N', \quad N' \ll N,
\end{equation}
that preserves the information necessary for downstream reasoning while reducing computational cost. Formally, given a language model $g(\cdot)$ operating on the token set with a text query $t$, we aim to ensure that

\begin{equation}
g(\mathcal{V}, t) \approx g(\mathcal{V}', t).
\end{equation}
meaning that the compressed representation maintains the semantic and spatial cues required for reasoning tasks while significantly reducing token cardinality.

\begin{figure}[t]
    \centering
    \includegraphics[width=1.0\linewidth]{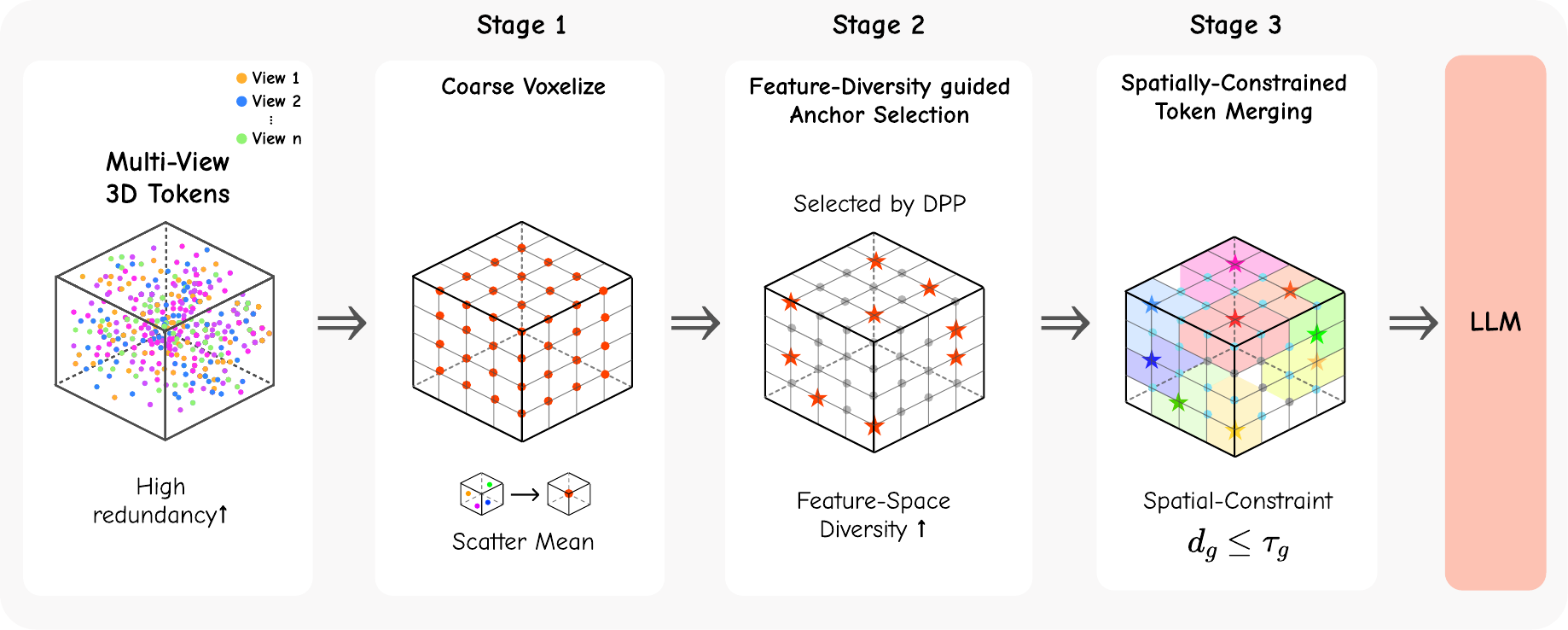}
    \caption{\textbf{Overview of the proposed three-stage token compression pipeline.} Given geometry-aware 3D tokens $\mathcal{V}$, we first apply coarse voxelization to obtain a spatially reduced set $\mathcal{V}_{\text{vox}}$. We then perform diversity-aware anchor selection via DPP on $\mathcal{V}_{\text{vox}}$ to identify semantically representative anchors $\mathcal{A}$. Finally, spatially-constrained token merging aggregates nearby non-anchor tokens into anchors to produce the compressed set $\mathcal{V}'$.}
    \label{fig:threestage}
    \vspace{-10pt}
\end{figure}

\vspace{-10pt}
\section{Method}
\vspace{-5pt}
Our goal is to compress the large set of geometry-aware 3D tokens $\mathcal{V}$ while preserving the semantic and spatial information required for downstream reasoning. As illustrated in Fig.~\ref{fig:threestage}, we design a three-stage token compression pipeline that progressively removes redundancy while maintaining geometric consistency.
Starting from the dense token set $\mathcal{V}$ generated by multi-view projection, we first apply \textbf{Stage 1: coarse voxelization} to reduce point-level redundancy caused by overlapping observations. Tokens within each voxel are aggregated to obtain a spatially reduced token set $\mathcal{V}_{\text{vox}}$.
Next, we perform \textbf{Stage 2: feature-diversity guided anchor selection}. From the voxelized token set $\mathcal{V}_{\text{vox}}$, we select a subset of representative anchors $\mathcal{A}$ using a Determinantal Point Process (DPP)~\cite{chen2018fast}, which encourages diversity in feature space and preserves semantically distinct tokens.
Finally, we apply \textbf{Stage 3: spatially-constrained token merging}. Non-anchor tokens are assigned to nearby anchors under a spatial distance constraint, allowing anchors to aggregate complementary contextual information while maintaining geometric consistency. This produces the compressed token set $\mathcal{V}'$, which is subsequently provided to the language model for downstream reasoning. The overall procedure of the proposed compression pipeline is summarized in Algorithm~\ref{alg:3dzip}.

\begin{algorithm}[t]
\caption{Overview of the proposed 3DZip}
\label{alg:3dzip}
\begin{algorithmic}[1]
\Require Original token set $\mathcal{V}=\{v_i\}_{i=1}^{N}$, voxel size $\delta$, anchor number $K$
\Ensure Compressed token set $\mathcal{V}'$

\Statex \textbf{Stage 1: Coarse Voxelization}
\State Partition tokens into voxel groups $\mathcal{G}=\{G_k\}$
\State Aggregate tokens within each voxel using mean pooling
\State Obtain voxel tokens $\mathcal{V}_{\text{vox}}=\{v_k\}$

\Statex \textbf{Stage 2: Feature-Diversity Guided Anchor Selection}
\State Normalize voxel features $\hat{\mathbf f}_k$
\State Construct similarity kernel $L_{kl} = \hat{\mathbf f}_k^\top \hat{\mathbf f}_l$
\State Select anchor set $\mathcal{A}$ via greedy DPP maximization with $|\mathcal{A}|=K$

\Statex \textbf{Stage 3: Spatially-Constrained Token Merging}
\For{each $j \in \mathcal{V}_{\text{vox}} \setminus \mathcal{A}$}
    \State Find nearest anchor $a^*(j)$ in feature space
    \If{$d_g(j,a^*(j)) \le \tau_g$}
        \State Assign token $j$ to anchor $a^*(j)$
    \Else
        \State Discard token $j$
    \EndIf
\EndFor
\State Aggregate assigned tokens into anchors

\State \Return Compressed token set $\mathcal{V}'$
\end{algorithmic}
\end{algorithm}
\vspace{-10pt}

\subsection{Stage 1: Coarse Voxelization}
\vspace{-5pt}
Multi-view projection often produces densely overlapping tokens corresponding to nearly identical 3D locations. To mitigate this \textit{point-level redundancy}, we partition the 3D space into axis-aligned voxels with size $\delta$.

Let $\mathcal{G}=\{G_k\}_{k=1}^{N_v}$ denote the set of occupied voxels, where each cell contains tokens whose positions fall inside the $k$-th voxel. Tokens within each voxel are aggregated via mean pooling

\begin{equation}
\mathbf f_k =
\frac{1}{|G_k|}
\sum_{v_i \in G_k}
\mathbf f_i,
\qquad
\mathbf p_k =
\frac{1}{|G_k|}
\sum_{v_i \in G_k}
\mathbf p_i .
\end{equation}

The resulting voxel tokens

\begin{equation}
v_k = (\mathbf f_k,\mathbf p_k)
\end{equation}

form the reduced set

\begin{equation}
\mathcal V_{\text{vox}} = \{v_k\}_{k=1}^{N_v}, \quad N_v \ll N .
\end{equation}

\vspace{-10pt}
\subsection{Stage 2: Feature-Diversity Guided Anchor Selection}

Although voxelization reduces point-level redundancy, many tokens still correspond to repeated observations of the same object surfaces, resulting in \textit{object-level redundancy}. To retain semantically distinct representations, we select a subset of representative anchors.

Given $\mathcal V_{\text{vox}}$, we aim to select an anchor set $\mathcal A \subset \mathcal V_{\text{vox}}$ with $|\mathcal A|=K$.  
We first normalize voxel features

\[
\hat{\mathbf f}_k = \frac{\mathbf f_k}{\|\mathbf f_k\|_2}.
\]

A cosine similarity kernel is then constructed

\begin{equation}
L_{kl} = \hat{\mathbf f}_k^\top \hat{\mathbf f}_l .
\end{equation}

Using a DPP, anchors are selected by maximizing $\det(L_{\mathcal A})$.  
We adopt the standard greedy approximation based on Cholesky decomposition~\cite{chen2018fast}.
\vspace{-5pt}
\subsection{Stage 3: Spatially-Constrained Token Merging}
\label{sec:method_stage3}
While anchors capture diverse semantic information, the remaining tokens may still 
contain complementary context. To incorporate such information while preserving 
spatial consistency, we merge nearby tokens into their assigned anchors.

For each non-anchor token index $j \in \mathcal V_{\text{vox}} \setminus \mathcal A$, 
we assign it to the most similar anchor in feature space,
\begin{equation}
a^*(j) =
\arg\min_{a \in \mathcal A}
\left(1 - \hat{\mathbf f}_j^\top \hat{\mathbf f}_a\right),
\end{equation}
and allow merging only when the spatial distance between tokens is sufficiently small.
Specifically, let
\begin{equation}
\mathbf c_k =
\left\lfloor \frac{\mathbf p_k}{\delta} \right\rfloor
\end{equation}
denote the voxel grid index of token $v_k$, and define the grid-space distance
\begin{equation}
d_g(j,a)=\|\mathbf c_j-\mathbf c_a\|_2 .
\end{equation}

We define the set of tokens assigned to anchor $a$ as
\begin{equation}
\mathcal S_a=
\{\, j \mid a^*(j)=a,\ d_g(j,a)\le\tau_g \,\}.
\end{equation}
Tokens for which no anchor satisfies the spatial constraint 
(i.e., $d_g(j, a^*(j)) > \tau_g$) are discarded, as they likely correspond 
to spatially isolated observations that do not contribute complementary context 
to any anchor. The anchor feature is then updated as the mean of itself and all assigned tokens:
\begin{equation}
\mathbf f'_a =
\frac{\mathbf f_a + \displaystyle\sum_{j\in\mathcal S_a}\mathbf f_j}
{1 + |\mathcal S_a|},
\qquad
\mathbf p'_a = \mathbf p_a .
\end{equation}

The final compressed token set is
\begin{equation}
\mathcal V' = \{v'_k\}_{k=1}^{N'},
\quad
v'_k = (\mathbf f'_k,\mathbf p'_k),
\quad
N' = |\mathcal A| = K \ll N .
\end{equation}
The resulting token set $\mathcal V'$ is then provided to the language model 
for downstream reasoning.
\vspace{-10pt}
\section{Experiments}
\vspace{-5pt}
\subsection{Experimental Settings}
\subsubsection{Baselines and Models}
We compare our method with representative token compression 
approaches, including voxelization-only aggregation and the 
voxel-based DTC. In addition, we evaluate methods originally 
developed for 2D VLM settings, including FastV~\cite{fastv}, 
SparseVLM~\cite{zhang2024sparsevlm}, VisionZip~\cite{yang2025visionzip}, 
and VisPruner~\cite{vispruner}, to examine their effectiveness 
when extended to 3D spatial representations. As the backbone 
model, we adopt LLaVA-3D~\cite{zhu2024llava} as our projection-based 3D
VLM. To ensure a fair comparison, all 
competing methods are implemented on top of the same backbone 
architecture.
\vspace{-10pt}
\subsubsection{Implementation Details}
We set the voxel size $\delta = 0.2$\,m for coarse 
voxelization and the grid-space threshold $\tau_g = 5$ 
for spatially-constrained merging. All experiments are 
conducted with temperature set to 0 to ensure 
reproducibility.
\vspace{-10pt}

\subsection{Dataset}
\label{sec:method_dataset}

\subsubsection{SQA3D}
This benchmark~\cite{ma2022sqa3d} extends 3D Question Answering to a 
situated and embodied setting, introducing 6.8K 
agent-centric situations over 650 ScanNet~\cite{dai2017scannet} scenes, with 
33K reasoning questions incorporating situation 
understanding and situated reasoning. We report 
exact match (EM) accuracy on the test set.
\vspace{-4mm}

\subsubsection{OpenEQA}
Proposed by~\cite{majumdar2024openeqa}, this is the first 
open-vocabulary benchmark for embodied reasoning in 3D 
environments, containing over 1.6K human-authored 
questions across seven reasoning categories from ScanNet 
and HM3D~\cite{ramakrishnan2021hm3d} environments. It employs an automated LLM-based 
evaluation pipeline, and we report the GPT-4o LLM-Match 
score following the standard protocol~\cite{hurst2024gpt4o}.
\vspace{-4mm}

\subsubsection{ScanQA}
This benchmark~\cite{azuma2022scanqa} is built on RGB-D 
reconstructions from ScanNet, containing 41K 
question–answer pairs over 800 indoor scenes. Questions 
require reasoning over object attributes, spatial 
relations, counting, and scene layout. We report EM and standard captioning metrics on the 
validation split.
\vspace{-4mm}

\begin{table}[t]
\centering
\caption{\textbf{Comparison of token compression methods.} The 3D-Aware column indicates whether spatial geometry is explicitly considered during token selection.}
\vspace{-5pt}
\setlength{\tabcolsep}{3pt} 
\renewcommand{\arraystretch}{0.5}
\resizebox{0.7\columnwidth}{!}{%
\begin{tabular}{lccccc}
\toprule
\textbf{Method} & \textbf{3D-Aware} & \textbf{ScanQA} & \textbf{SQA3D} & \textbf{OpenEQA} & \textbf{Rel.} \\
\midrule

\rowcolor{cyan!8}
\multicolumn{6}{c}{\emph{All 1410 Tokens}} \\
LLaVA-3D (ICCV'25) & \checkmark & 26.5 & 55.7 & 60.3 & 100.0\% \\
\midrule

\rowcolor{cyan!8}
\multicolumn{6}{c}{\emph{Retain 128 Tokens ↓(9.1\%)}} \\
FastV (ECCV'24) & \ding{55} & 21.9 & 50.9 & 56.0 & 88.9\% \\
SparseVLM (ICML'25) & \ding{55} & 21.8 & 50.3 & 53.9 & 87.3\% \\
VisionZip (CVPR'25) & \ding{55} & 22.2 & 51.5 & 56.2 & 89.8\% \\
VisPruner (ICCV'25) & \ding{55} & 22.3 & 52.0 & 55.8 & 90.0\% \\
\cmidrule(lr){1-6}
Voxelization & \checkmark & 23.6 & 51.5 & 54.6 & 90.7\% \\
DTC (CVPR'25) & \checkmark & 22.1 & 51.3 &  54.8 & 88.8\% \\
\textbf{3DZip (Ours)} & \checkmark &\textbf{24.2} & \textbf{53.2} & \textbf{58.6} & \textbf{94.7\%} \\
\midrule

\rowcolor{cyan!8}
\multicolumn{6}{c}{\emph{Retain 64 Tokens ↓(4.5\%)}} \\
FastV (ECCV'24) & \ding{55} & 21.1 & 49.5 & 53.8 & 85.9\% \\
SparseVLM (ICML'25) & \ding{55} & 20.9 & 49.8 & 53.6 & 85.7\% \\
VisionZip (CVPR'25) & \ding{55} & 20.0 & 49.1 & 53.4 & 84.1\% \\
VisPruner (ICCV'25) & \ding{55} & 21.9  & 50.1 & 53.9 & 87.3\% \\
\cmidrule(lr){1-6}
Voxelization & \checkmark  & 21.9 & 49.8 & 54.6 & 87.5\% \\
DTC (CVPR'25) & \checkmark & 21.3 & 50.2 &  54.3 & 86.8\% \\

\textbf{3DZip (Ours)} & \checkmark & \textbf{23.3} & \textbf{52.8} &  \textbf{56.7} & \textbf{92.3\%} \\
\midrule

\rowcolor{cyan!8}
\multicolumn{6}{c}{\emph{Retain 32 Tokens ↓(2.3\%)}} \\
FastV (ECCV'24) & \ding{55} & 19.9 & 47.7 & 52.5 & 82.6\% \\
SparseVLM (ICML'25) & \ding{55} & 20.4 & 48.7 & 52.2 & 83.7\% \\
VisionZip (CVPR'25) & \ding{55} & 19.6 & 47.0 & 52.0 & 81.5\% \\
VisPruner (ICCV'25) & \ding{55} & 20.9 & 49.1 & 53.5& 85.2\% \\
\cmidrule(lr){1-6}
Voxelization & \checkmark & 20.7 & 47.9 & 53.4 & 84.2\% \\
DTC (CVPR'25) & \checkmark & 20.6 & 48.8 & 52.6 & 84.2\% \\
\textbf{3DZip (Ours)}  & \checkmark & \textbf{21.9} & \textbf{51.1} &  \textbf{55.7} & \textbf{88.9\%} \\
\bottomrule
\end{tabular}%
}

\vspace{-15pt}
\label{tab:main_results}
\end{table}

\begin{table*}[t]
\centering
\caption{\textbf{Category-wise OpenEQA results.} 3DZip achieves the highest average at all token budgets, 
with the largest gains in attribute recognition and object recognition.}
\small
\resizebox{\textwidth}{!}{
\begin{tabular}{lcccccccc}
\toprule
& \multicolumn{8}{c}{\textbf{OpenEQA}} \\
\cmidrule(lr){2-8}
\textbf{Method}
& \makecell{\textbf{attribute}\\\textbf{recognition}}
& \makecell{\textbf{functional}\\\textbf{reasoning}}
& \makecell{\textbf{object}\\\textbf{localization}}
& \makecell{\textbf{object}\\\textbf{recognition}}
& \makecell{\textbf{object state}\\\textbf{recognition}}
& \makecell{\textbf{spatial}\\\textbf{understanding}}
& \makecell{\textbf{world}\\\textbf{knowledge}}
& \makecell{\textbf{Avg.}} \\
\midrule
\rowcolor{cyan!8}
\multicolumn{9}{c}{\emph{All 1410 Tokens}} \\
LLaVA-3D (ICCV'25)
& 64.0
& 62.7
& 51.5
& 55.9
& 73.7
& 54.2
& 59.1  
& 60.3 \\
\midrule
\rowcolor{cyan!8}
\multicolumn{9}{c}{\emph{Retain 128 Tokens ↓(9.1\%)}} \\
FastV(ECCV'24)
& 57.8
& \underline{58.9}
& 46.8
& 47.7
& \textbf{75.6}
& 45.3
& 59.4
& 56.0 \\
SparseVLM(ICML'25)
& 54.0
& 54.1
& 48.3
& 45.9
& \underline{74.1}
& 44.5
& 55.5
& 53.9 \\

VisionZip(CVPR'25)
& \underline{58.4}
& 57.8
& 48.4
& \underline{51.0}
& 69.4
& 47.2
& \textbf{60.9}
& \underline{56.2} \\

Voxelization
& 56.3
& 58.5
& 47.2
& 47.2
& 68.6
& \underline{49.1}
& 54.9
& 54.6 \\

DTC(CVPR'25)
& 56.4
& 57.7
& \underline{48.5}
& 46.2
& 71.5
& 47.2
& 55.3
& 54.8 \\

\textbf{3DZip (Ours)}
& \textbf{64.2}
& \textbf{60.0}
& \textbf{50.0}
& \textbf{53.2}
& 72.5
& \textbf{49.5}
& \underline{60.1}
& \textbf{58.6} \\

\midrule
\rowcolor{cyan!8}
\multicolumn{9}{c}{\emph{Retain 64 Tokens ↓(4.5\%)}} \\
FastV(ECCV'24)
& 51.2
& 58.2
& 47.1  
& 43.2
& \textbf{76.3}
& 43.2
& \underline{56.7}
& 53.8 \\
SparseVLM(ICML'25)
& 50.6
& 57.8
& 46.5
& 45.3
& \underline{74.3}
& 45.2
& 55.5
& 53.6 \\

VisionZip(CVPR'25)
& 51.7
& 57.4
& \underline{48.6}
& 44.7
& 69.4
& 45.6
& 56.3
& 53.4 \\

Voxelization
& 50.8
& \textbf{61.4}
& 46.2
& \underline{47.9}
& 71.5
& \textbf{48.5}
& 55.7
& \underline{54.6} \\

DTC(CVPR'25)
& \underline{55.5}
& 58.2  
& 45.5
& 43.2
& 74.1
& 47.8
& 55.0
& 54.3 \\

\textbf{3DZip (Ours)}
& \textbf{60.9}
& \underline{59.7}
& \textbf{48.8}
& \textbf{51.2}
& 69.6
& \underline{48.1}
& \textbf{58.0}
& \textbf{56.7} \\

\midrule
\rowcolor{cyan!8}
\multicolumn{9}{c}{\emph{Retain 32 Tokens ↓(2.3\%)}} \\
FastV(ECCV'24)
& 50.7
& 58.6
& 43.9
& 41.2
& \textbf{73.6}
& 42.6
& 56.6
& 52.5 \\
SparseVLM(ICML'25)
& 50.8
& 57.4
& 45.6
& 40.6
& \underline{73.3}
& 42.9
& 54.2
& 52.2 \\
VisionZip(CVPR'25)
& 51.0
& 57.0
& 45.2
& 42.6
& 68.1
& 44.4
& 55.6
& 52.0 \\

Voxelization
& 50.4
& \textbf{60.3}
& \underline{46.9}
& \underline{43.8}
& 68.8
& \underline{46.1}
& \textbf{57.7}
& \underline{53.4} \\

DTC(CVPR'25)
& \underline{51.3}
& 56.8
& 45.1
& 42.9
& 72.2
& 44.2
& 55.6
& 52.6 \\

\textbf{3DZip (Ours)}
& \textbf{56.3}
& \underline{59.7}
& \textbf{47.4}
& \textbf{50.5}
& 70.7 
& \textbf{47.7}
& \underline{57.4}
& \textbf{55.7} \\
\bottomrule
\end{tabular}
}
\vspace{-15pt}
\label{tab:eqa_category}
\end{table*}
\subsection{Results}
\subsubsection{Performance Comparison} Table~\ref{tab:main_results} compares our method with representative token compression approaches under different token budgets on ScanQA, SQA3D, and OpenEQA.  Methods originally designed for 2D VLMs, such as FastV, SparseVLM, VisionZip, and VisPruner, consistently underperform in 3D Question Answering settings. In particular, VisionZip and VisPruner select tokens based on CLS-attention from the vision encoder, emphasizing global semantic relevance. While effective for 2D tasks, such global attention-based selection does not explicitly preserve local geometric relationships required for spatial reasoning in 3D Question Answering. Similarly, LLM-attention-based pruning selects tokens with high relevance to the input text, which can bias the selection toward a small set of semantically salient regions while discarding surrounding tokens that encode spatial context, potentially disrupting geometric coherence in 3D scenes.

However, existing 3D-aware methods still show limited performance under constrained token budgets. Purely spatial aggregation, such as voxelization, may overlook semantic representativeness. Although DTC incorporates visual semantics, it still shows performance degradation as the token budget becomes more constrained.
Across all token budgets (128, 64, and 32 tokens), our method consistently achieves the best performance. Notably, 3DZip maintains the highest relative performance (Rel.), preserving 94.7\% and 92.3\% of uncompressed performance at 128 and 64 tokens, respectively, consistently outperforming the best baselines. For example, at 64 tokens, our method improves SQA3D from 50.2 (DTC) and 49.1 (VisionZip) to 52.8, and even under aggressive compression (32 tokens), it maintains strong performance (51.1), outperforming both 2D-based and existing 3D-aware baselines. These results indicate that jointly preserving feature-space diversity and geometric coherence is crucial for effective 3D token compression. Supp. Sec.~\ref{sec:other_backbone} further reports 
results on additional backbones (Video-3D-LLM~\cite{zheng2025video3d} 
and SR-3D~\cite{sr3d}), and Supp. Sec.~\ref{sec:other_metric} reports 
additional ScanQA metrics, confirming the consistent gains of 3DZip.

\vspace{-4mm}
\subsubsection{Category-wise Analysis on OpenEQA}

Table~\ref{tab:eqa_category} presents a category-wise breakdown on OpenEQA across seven capabilities. 3DZip achieves the highest average score at all token budgets. In \textit{attribute recognition}, 3DZip scores 64.2 at 128 tokens, closely matching the uncompressed baseline (64.0) and outperforming the second-best method by +5.8 points. In \textit{object recognition}, it retains 53.2 at 128 tokens and maintains 50.5 even at 32 tokens, yielding the largest margin (+6.7) at this budget.

3DZip also consistently improves performance in \textit{object localization} and \textit{spatial understanding}, the two most spatially demanding categories. Methods such as FastV and SparseVLM rely on LLM attention scores, which tend to emphasize globally salient regions while overlooking spatially distributed tokens that encode positional cues. VisionZip mitigates this through CLS-attention-based selection with token merging, and DTC applies spatial aggregation, but both rely on averaging operations that may dilute fine-grained geometric details. In contrast, 3DZip's DPP-based selection explicitly promotes feature diversity, helping preserve visually distinct and spatially informative tokens.

Furthermore, tasks reliant on prior knowledge and commonsense, such as \textit{functional reasoning} and \textit{world knowledge} prove robust to aggressive compression, with 3DZip maintaining stable performance across all budgets. An exception is \textit{object state recognition}, where FastV slightly exceeds the uncompressed baseline (75.6 vs.\ 73.7). This category yields consistently high scores across all methods, indicating it is inherently easier—primarily involving binary-style judgments (e.g., open vs.\ closed) that require minimal spatial reasoning or geometric understanding of inter-object relationships. Moreover, since FastV and SparseVLM select tokens based on LLM attention guided by text-query relevance, they naturally excel at retaining tokens specific to such narrowly focused, object-centric state queries.

\begin{figure}[t]
    \centering
    \includegraphics[width=0.86\linewidth]{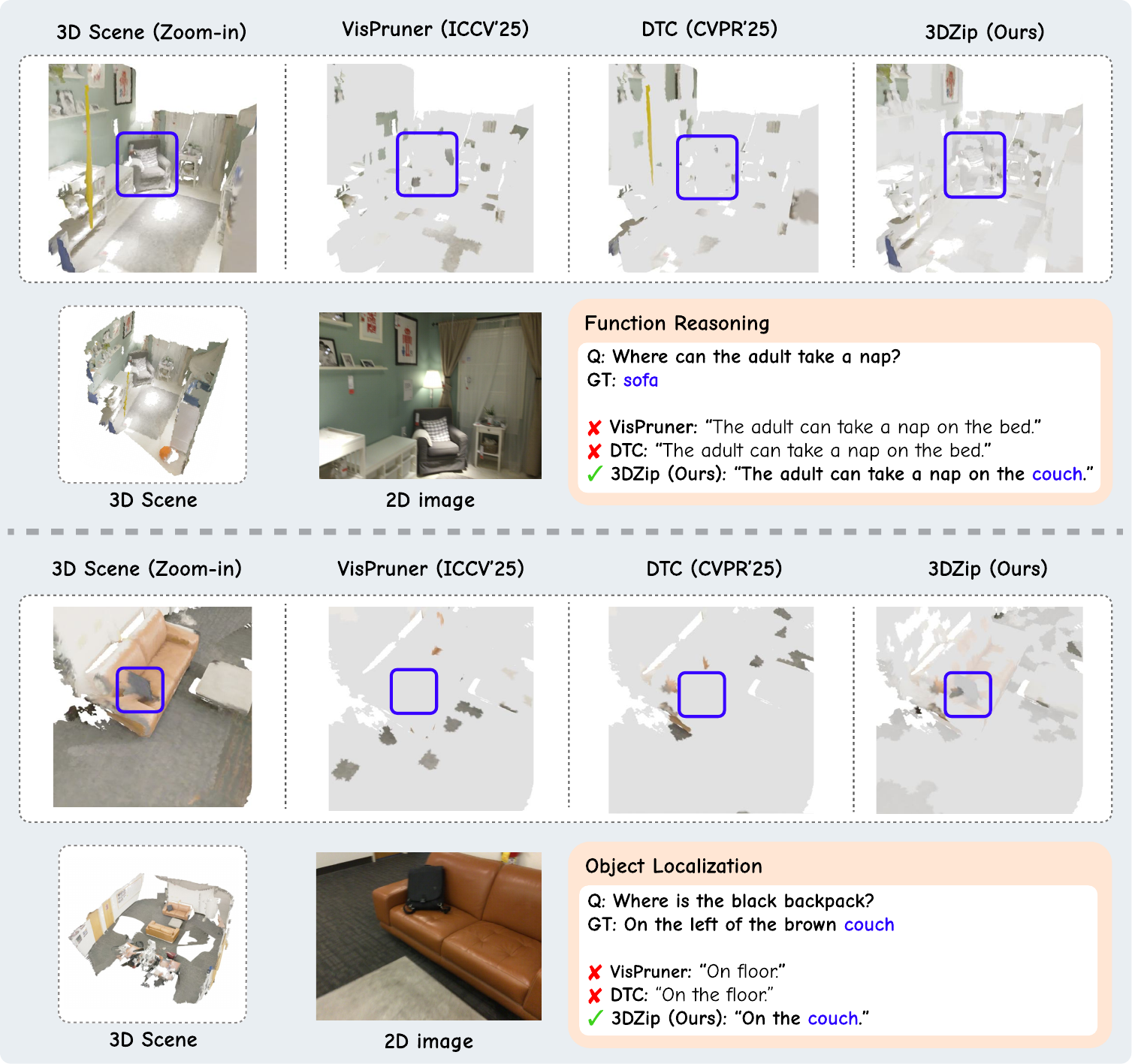}
    \caption{\textbf{Qualitative comparison of retained 3D tokens on OpenEQA.} 
VisPruner tends to allocate many tokens to floor regions, while DTC may miss some key objects in the scene. In contrast, 3DZip retains tokens on multiple semantically important objects and further enriches them through spatially-constrained merging.}
    \label{fig:qualitative}
    \vspace{-15pt}
\end{figure}

\subsubsection{Qualitative Results}
Fig.~\ref{fig:qualitative} visualizes the preserved 3D tokens and corresponding question answering results. 2D-attention-based methods like VisPruner tend to allocate many tokens to background and floor regions, missing critical local objects. Methods like DTC, which perform token matching within individual voxels, may struggle to capture object-level diversity across the entire scene, potentially leaving room for a relative concentration of tokens on large uniform structures. In contrast, 3DZip captures object-level diversity across the entire scene through feature-space DPP-based anchor selection, and further enriches each anchor with complementary context through spatially-constrained merging.

\begin{table*}[t]
    \centering
    \caption{\textbf{Component-wise ablation results on SQA3D.} (a) feature vs. spatial distance in anchor selection, (b) the effect of varying voxel sizes ($\delta$) in coarse voxelization, and (c) the role of the merging step and spatial constraints.}
    \resizebox{0.8\columnwidth}{!}{%
    \begin{subtable}[t]{0.32\textwidth}
        \centering
        \begin{tabular*}{\linewidth}{@{\extracolsep{\fill}}lc}
        \toprule
        \textbf{Method} & \textbf{EM} \\
        \midrule
        \rowcolor{gray!8}
        \multicolumn{2}{c}{\emph{All 1410 Tokens}} \\
        LLaVA-3D & 55.7 \\
        \midrule
        \rowcolor{gray!8}
        \multicolumn{2}{c}{\emph{Retain 64 Tokens}} \\
        Voxelization & 49.8 \\
        DTC (CVPR'25) & 50.2 \\
        Spatial distance & 50.1 \\
        \textbf{Feature distance} & \textbf{52.8} \\
        \midrule
        \rowcolor{gray!8}
        \multicolumn{2}{c}{\emph{Retain 32 Tokens}} \\
        Voxelization & 47.9 \\
        DTC (CVPR'25) & 48.8 \\
        Spatial distance & 49.0 \\
        \textbf{Feature distance} & \textbf{51.1} \\
        \bottomrule
        \end{tabular*}
        \subcaption{}
        \label{tab:3a}
    \end{subtable}
    \hspace{0.015\textwidth}
    \begin{subtable}[t]{0.32\textwidth}
        \centering
        \begin{tabular*}{\linewidth}{@{\extracolsep{\fill}}lc}
        \toprule
        \textbf{Method} & \textbf{EM} \\
        \midrule
        \rowcolor{gray!8}
        \multicolumn{2}{c}{\emph{All 1410 Tokens}} \\
        LLaVA-3D & 55.7 \\
        \midrule
        \rowcolor{gray!8}
        \multicolumn{2}{c}{\emph{Retain 64 Tokens}} \\
        w/o Coarse voxelize & 51.8 \\
        $\delta$=0.1m & 52.6 \\
        \textbf{$\boldsymbol{\delta}$=0.2m} & \textbf{52.8} \\
        $\delta$=0.3m & 52.1 \\
        \midrule
        \rowcolor{gray!8}
        \multicolumn{2}{c}{\emph{Retain 32 Tokens}} \\
        w/o Coarse voxelize & 50.2 \\
        $\delta$=0.1m & 50.9 \\
        \textbf{$\boldsymbol{\delta}$=0.2m} & \textbf{51.1} \\
        $\delta$=0.3m & 50.6 \\
        \bottomrule
        \end{tabular*}
        \subcaption{}
        \label{tab:ablation_stage1}
    \end{subtable}
    \hspace{0.015\textwidth}
    \begin{subtable}[t]{0.32\textwidth}
        \centering
        \begin{tabular*}{\linewidth}{@{\extracolsep{\fill}}lc}
        \toprule
        \textbf{Method} & \textbf{EM} \\
        \midrule
        \rowcolor{gray!8}
        \multicolumn{2}{c}{\emph{All 1410 Tokens}} \\
        LLaVA-3D & 55.7 \\
        \midrule
        \rowcolor{gray!8}
        \multicolumn{2}{c}{\emph{Retain 64 Tokens}} \\
        DTC (CVPR'25) & 50.2 \\
        w/o Merge & 52.5 \\
        w/o Spatial-Const. & 52.3 \\
        \textbf{w/ All} & \textbf{52.8} \\
        \midrule
        \rowcolor{gray!8}
        \multicolumn{2}{c}{\emph{Retain 32 Tokens}} \\
        DTC (CVPR'25) & 48.8 \\
        w/o Merge & 50.5 \\
        w/o Spatial-Const. & 50.8 \\
        \textbf{w/ All} & \textbf{51.1} \\
        \bottomrule
        \end{tabular*}
        \subcaption{}
        \label{tab:3c}
    \end{subtable}
    }
    \label{tab:main_ablation}
    \vspace{-20pt}
\end{table*}

\vspace{-10pt}
\subsection{Ablations}

\subsubsection{Feature Distance vs. Spatial Distance}
Table~\ref{tab:3a} compares feature-space distance and spatial (XYZ) distance as the kernel used in DPP-based anchor selection. Across all token budgets, feature-space distance consistently outperforms spatial distance by a clear margin (e.g., 52.8 vs. 50.1 EM at 64 tokens). When spatial distance is used, DPP favors geometrically dispersed anchors, promoting broad spatial coverage. However, spatially distant regions may still contain visually similar or redundant content—for example, multiple anchors may be redundantly allocated to different parts of a large wall or floor. Such spatially biased selection leads to a long-tail 
distribution in token allocation, where tokens are disproportionately concentrated on a few large objects, hindering overall object coverage across the scene.

In contrast, feature-space distance directly captures semantic dissimilarity, encouraging the selection of tokens that represent distinct objects rather than repeated surface observations. This mitigates the long-tail bias in token allocation and improves object coverage across the scene. These results support our hypothesis that feature dispersion, rather than geometric dispersion, is the more effective criterion for 3D token compression. Beyond this comparison, Supp. Sec.~\ref{sec:diversity_comparison} and~\ref{sec:distance_metric} show that alternative diversity-based anchor selection methods are also effective, and that cosine similarity performs best among feature-space metrics.

\vspace{-4mm}
\subsubsection{Ablation on Coarse Voxelize}
Table~\ref{tab:ablation_stage1} investigates the impact of the initial coarse voxelization step and the choice of voxel size $\delta$. Removing the voxelization stage entirely (\emph{w/o Coarse Voxelize}) leads to a noticeable performance drop (e.g., from 52.8 to 51.8 EM under the 64-token budget), confirming that the severe multi-view redundancy must be resolved before feature-diversity-based anchor selection. Without this stage, the DPP anchor selection operates purely in feature space, which can result in clustered anchors that fail to preserve the overall 3D geometry of the scene. However, overly coarse voxelization is also harmful: as shown with $\delta = 0.3$m, a larger voxel size prematurely merges distinct local objects, destroying fine-grained spatial cues essential for 3D reasoning. These results highlight the importance of balancing multi-view redundancy reduction and local geometric preservation when compressing 3D tokens. To complement this ablation, Supp. Sec.~\ref{sec:point_level_redunduncy} quantitatively shows that coarse voxelization reduces point-level redundancy caused by duplicate multi-view observations.

\vspace{-4mm}
\subsubsection{Effect of Spatially-Constrained Token Merging}
Table~\ref{tab:3c} ablates the merging strategy and spatial constraints under different token budgets. Removing the merging step (\emph{w/o Merge}) leads to a performance drop (e.g., from 52.8 to 52.5 EM at 64 tokens), confirming that aggregating non-anchor tokens into nearby anchors enriches anchor representations with complementary context. Omitting spatial constraints (\emph{w/o Spatial-Const.}) results in a similar decrease (e.g., from 52.8 to 52.3 at 64 tokens), indicating that geometrically constrained merging is essential for maintaining structural consistency. Combining both components consistently yields the best performance across all compression ratios. Additional results on the spatial threshold $\tau_g$ are provided in Supp. Sec.~\ref{sec:spatial_constraints}.

\begin{table}[t]
\centering
\caption{Efficiency and accuracy comparison on a single RTX 4090 using SQA3D.}
\small
\resizebox{0.75\columnwidth}{!}{%
\begin{tabular*}{\textwidth}{@{\extracolsep{\fill}} l c c c c c}
\toprule
\textbf{\shortstack{\raisebox{0.5ex}{Method}}} &
\textbf{\shortstack{Retain\\Tokens}} & 
\textbf{\shortstack{FLOPs\\(T)}} & 
\textbf{\shortstack{Latency\\(ms/sample)}} & 
\textbf{\shortstack{Cache Size\\(MB)}} & 
\textbf{\raisebox{0.5ex}{EM}} \\
\midrule
LLaVA-3D & 1410 & 9.18  & 342  & 722 & 55.7 \\
\midrule
FastV (ECCV'24) & 128 & 1.41 & 191 & 139 & 50.9 \\
DTC (CVPR'25) & 128 & 0.90 & 196 & 	101 & 51.3 \\
\textbf{3DZip (Ours)} & 128 &  \textbf{0.90}& \textbf{178} & \textbf{101}& \textbf{53.2} \\
\bottomrule
\end{tabular*}
}
\vspace{-15pt}
\label{tab:latency_throughput}
\end{table}

\vspace{-4mm}
\subsection{Efficiency Analysis}
\label{sec:Efficiency}
\vspace{-2pt}
We evaluate computational efficiency on a single NVIDIA RTX 4090 GPU using the SQA3D benchmark under identical settings. As shown in Table~\ref{tab:latency_throughput}, our method reduces inference latency from 342ms/sample (LLaVA-3D) to 178ms/sample, achieving a 48.0\% reduction. Under the same 128-token budget, our method outperforms DTC in practical speed (196ms vs. 178ms), yielding a 9.2\% speedup. This efficiency gain stems from aggressive yet effective token reduction; compared to the LLaVA-3D baseline (1410 tokens), our method retains only 128 tokens ($-$90.9\%), which reduces FLOPs from 9.18T to 0.90T ($-$90.2\%) and KV cache size from 722MB to 101MB ($-$86.0\%). Despite these substantial reductions, the EM score drops by only 2.5 points (55.7 to 53.2), demonstrating a highly favorable efficiency--accuracy trade-off. Notably, while DTC achieves similar theoretical FLOPs for a given token count, our method is more efficient in practice due to our one-pass geometry-aware selection, which avoids the overhead of iterative matching and refinement. Furthermore, by removing redundant tokens before the LLM backbone, we achieve significant savings in both attention computation and KV cache footprint. Our implementation is also compatible with FlashAttention~\cite{dao2022flashattention} for further acceleration.

\section{Limitations and Future Work}
While 3DZip achieves a strong efficiency--accuracy trade-off, it still has several limitations. The coarse voxelization in Stage 1 may attenuate fine-grained cues for small objects, since their tokens can be merged with nearby objects or background regions within the same voxel, as analyzed in Supp. Sec. ~\ref{sec:finegrained}. In addition, the hyperparameters $(\delta, \tau_g)$ are fixed across scenes rather than adapted to scene complexity. Since scene size can affect object density and inter-object distances, fixed hyperparameters may be suboptimal across different environments. Although their sensitivity to scene scale is examined in Supp. Sec.~\ref{sec:scenesize}, developing adaptive hyperparameter selection that accounts for scene complexity and small-object density remains a promising future direction.

\section{Conclusion}
In this work, we present 3DZip, a geometry-aware token compression framework for projection-based 3D VLMs. We identify two distinct forms of redundancy inherent to multi-view 3D representations—point-level redundancy caused by overlapping observations and object-level redundancy arising from repeated surface tokens of the same object—and demonstrate that spatial aggregation alone is insufficient to address the latter. To this end, 3DZip combines three complementary stages: coarse voxelization to reduce point-level redundancy, feature-diversity-guided anchor selection via DPP to capture semantically distinct objects across the scene, and spatially-constrained token merging to enrich anchors with complementary context while preserving geometric consistency. Extensive experiments on ScanQA, SQA3D, and OpenEQA demonstrate that 3DZip consistently outperforms both 2D-based and existing 3D-aware token compression methods across all token budgets, retaining 94.7\% of the original accuracy with only 128 tokens and achieving a $1.92\times$ speedup. We hope that our analysis of object-level token imbalance and the proposed feature-diversity-driven compression strategy provide useful insights for future work on efficient 3D VLMs.

\section*{Acknowledgements}
\vspace{-5pt}
This work was supported by the National Research Foundation of Korea(NRF) grant funded by the Korean government(MSIT) (No. RS-2024-00456152) and the ``Advanced GPU Utilization Support Program” funded by the Government of the Republic of Korea(Ministry of Science and ICT), and the authors gratefully acknowledge the Cluster Server for Computational Science at Pusan National University for providing computational resources.

\bibliographystyle{splncs04}
\bibliography{main}

\appendix
\clearpage

\makeatletter
\renewcommand{\theHsection}{appendix.\Alph{section}}
\renewcommand{\theHsubsection}{appendix.\Alph{section}.\arabic{subsection}}
\renewcommand{\theHsubsubsection}{appendix.\Alph{section}.\arabic{subsection}.\arabic{subsubsection}}
\makeatother

\section*{Supplementary Material Overview}
This supplementary material provides additional experimental results, analyses, and implementation details supporting the main paper.
\begin{itemize}[itemsep=4pt, topsep=6pt]

\item \textbf{A. Additional Results}
\begin{itemize}[itemsep=3pt]
\item \ref{sec:other_backbone} Evaluation on Additional Models
\item \ref{sec:3dcap} Evaluation on 3D Dense Captioning
\item \ref{sec:other_metric} Additional Metrics
\item \ref{sec:qualitative} Additional Qualitative Results
\end{itemize}

\item \textbf{B. Additional Analysis}
\begin{itemize}[itemsep=3pt]
\item \ref{sec:point_level_redunduncy} Point-Level Token Redundancy Analysis
\item \ref{sec:object_coverage} Object-Level Token Allocation Analysis
\item \ref{sec:diversity_comparison} Comparison of Diversity Strategies
\item \ref{sec:distance_metric} Distance Metric Analysis
\item \ref{sec:spatial_constraints} Effect of Spatial Constraints
\item \ref{sec:divprune} Comparison with 2D Diversity-based Token Compression
\item \ref{sec:stage2_robustness} Robustness of Stage 2 to Stage-1 Aggregation Strategies
\item \ref{sec:finegrained} Fine-grained Analysis and Failure Cases
\item \ref{sec:scenesize} Effect of Scene Size on Hyperparameters
\end{itemize}

\item \textbf{C. Method Details}
\begin{itemize}[itemsep=3pt]
\item \ref{sec:dpp_algo} DPP Algorithm
\item \ref{sec:dpp_complexity} DPP Runtime Analysis
\end{itemize}

\item \textbf{D. Experimental Details}
\begin{itemize}[itemsep=3pt]
\item \ref{sec:sup_exp_details} Experimental Setup
\item \ref{sec:dataset_details} Dataset Details
\item \ref{sec:llm_eval} LLM-based Evaluation
\end{itemize}

\end{itemize}

\newpage

\begin{table*}[t]
\caption{\textbf{Generalization across projection-based 3D VLM models.} 
We evaluate 3DZip on two recent projection-based 3D VLMs, Video-3D-LLM and SR-3D. 
Under identical token budgets, 3DZip consistently outperforms all baselines across both architectures, 
demonstrating that the proposed compression strategy generalizes beyond the LLaVA-3D model used in the main experiments.}
\centering
\begin{subtable}[t]{0.48\textwidth}
    \centering
    \setlength{\tabcolsep}{3pt}
    \renewcommand{\arraystretch}{0.87}
    \resizebox{\columnwidth}{!}{%
    \begin{tabular}{lccc}
    \toprule
    \textbf{Method} & \textbf{ScanQA} & \textbf{SQA3D} & \textbf{OpenEQA} \\
    \midrule
    \rowcolor{orange!8}
    \multicolumn{4}{c}{\emph{All 3920 Tokens}} \\
    Video-3D-LLM (CVPR'25) & 29.9 & 58.4 & 59.4 \\
    \midrule

    \rowcolor{orange!8}
    \multicolumn{4}{c}{\emph{Retain 128 Tokens ↓(3.3\%)}} \\
    VisPruner (ICCV'25) & 23.6 & 51.2 & 53.5 \\
    Voxelization        & 24.6 & 52.2 & 52.9 \\
    DTC (CVPR'25)       & 23.7 & 51.9 & 53.5 \\
    \textbf{3DZip (Ours)} & \textbf{24.7} & \textbf{53.1} & \textbf{54.6} \\
    \midrule

    \rowcolor{orange!8}
    \multicolumn{4}{c}{\emph{Retain 64 Tokens ↓(1.6\%)}} \\
    VisPruner (ICCV'25) & 22.3 & 49.2 & 51.4 \\
    Voxelization        & 22.7 & 50.0 & 51.9 \\
    DTC (CVPR'25)       & 22.5 & 49.2 & 52.0 \\
    \textbf{3DZip (Ours)} & \textbf{23.8} & \textbf{52.2} & \textbf{54.6} \\
    \midrule

    \rowcolor{orange!8}
    \multicolumn{4}{c}{\emph{Retain 32 Tokens ↓(0.8\%)}} \\
    VisPruner (ICCV'25) & 21.0 & 48.4 & 50.2 \\
    Voxelization        & 20.9 & 48.3 & 50.3 \\
    DTC (CVPR'25)       & 20.8 & 49.0 & 50.0 \\
    \textbf{3DZip (Ours)} & \textbf{23.3} & \textbf{51.8} & \textbf{53.0} \\
    \bottomrule
    \end{tabular}%
    }
    \subcaption{\textbf{Video-3D-LLM}~\cite{zheng2025video3d}}
    \label{tab:video3dllm}
\end{subtable}
\hfill
\begin{subtable}[t]{0.48\textwidth}
    \centering
    \setlength{\tabcolsep}{3pt}
    \renewcommand{\arraystretch}{0.6}
    \resizebox{\columnwidth}{!}{%
    \begin{tabular}{lccc}
    \toprule
    \textbf{Method} & \textbf{ScanQA} & \textbf{SQA3D} & \textbf{OpenEQA} \\
    \midrule
    \rowcolor{orange!8}
    \multicolumn{4}{c}{\emph{All 2420 Tokens}} \\
    SR-3D (ICLR'26) & 29.6 & 61.0 & 61.7 \\
    \midrule
    \rowcolor{orange!8}
    \multicolumn{4}{c}{\emph{Retain 128 Tokens ↓(5.3\%)}} \\
    VisPruner (ICCV'25) & 23.0 & 51.5 & 54.3 \\
    Voxelization        & 22.8 & 51.1 & 52.2 \\
    DTC (CVPR'25)       & 23.5 & 51.4 & 53.5 \\
    \textbf{3DZip (Ours)} & \textbf{24.2} & \textbf{53.2} & \textbf{54.5} \\
    \midrule
    \rowcolor{orange!8}
    \multicolumn{4}{c}{\emph{Retain 64 Tokens ↓(2.6\%)}} \\
    VisPruner (ICCV'25) & 21.9 & 50.3 & 52.0 \\
    Voxelization        & 21.1 & 48.8 & 49.6 \\
    DTC (CVPR'25)       & 21.1 & 49.6 & 52.0 \\
    \textbf{3DZip (Ours)} & \textbf{23.1} & \textbf{51.9} & \textbf{52.6} \\
    \midrule
    \rowcolor{orange!8}
    \multicolumn{4}{c}{\emph{Retain 32 Tokens ↓(1.3\%)}} \\
    VisPruner (ICCV'25) & 20.1 & 49.1 & 51.5 \\
    Voxelization        & 19.9 & 47.9 & 50.3 \\
    DTC (CVPR'25)       & 19.7 & 48.1 & 51.4 \\
    \textbf{3DZip (Ours)} & \textbf{21.5} & \textbf{49.6} & \textbf{51.7} \\
    \bottomrule
    \end{tabular}%
    }
    \subcaption{\textbf{SR-3D}~\cite{sr3d}}
    \label{tab:sr3d}
\end{subtable}

\vspace{-10pt}
\label{tab:additional_backbone}
\end{table*}

\vspace{-10pt}
\section{Additional Results}
\label{sec:additional_results}
\vspace{-5pt}
\subsection{Evaluation on Additional Models}
\label{sec:other_backbone}

To verify the generalizability of 3DZip beyond LLaVA-3D, we evaluate our method on two recent projection-based 3D VLMs, Video-3D-LLM~\cite{zheng2025video3d} and SR-3D~\cite{sr3d}, both using 20 input views per scene. These models differ substantially in their visual encoding pipelines and token construction strategies, providing a strong testbed for evaluating the robustness of the proposed compression framework.
For a fair comparison, 3DZip is applied after the 3D token construction stage in each backbone, without modifying the original model architecture or training procedure. This ensures that the compression strategy operates under identical settings across models.
As shown in Table~\ref{tab:additional_backbone}, 3DZip consistently outperforms all baselines across all token budgets on both models. The improvements are particularly noticeable under aggressive compression (e.g., 64 and 32 tokens), where preserving semantically diverse tokens becomes critical. These results demonstrate that the proposed compression strategy is not tied to a specific architecture, but instead generalizes well across different projection-based 3D VLM designs.

\begin{table}[t]
\caption{\textbf{3D dense captioning results on the Scan2Cap benchmark.}
We compare 3DZip with existing token compression methods under different token budgets.}
\centering
\renewcommand{\arraystretch}{0.6}
\resizebox{0.65\columnwidth}{!}{%
\begin{tabular}{lcccc}
\toprule
\textbf{Method} & \textbf{CIDEr} & \textbf{BLEU-4} & \textbf{METEOR} & \textbf{ROUGE-L} \\
\midrule

\rowcolor{orange!8}
\multicolumn{5}{c}{\emph{All 1410 Tokens (Baseline)}} \\
LLaVA-3D (ICCV'25) & 65.8 & 11.2 & 15.5 & 37.0 \\
\midrule

\rowcolor{orange!8}
\multicolumn{5}{c}{\emph{Retain 128 Tokens ↓(9.1\%)}} \\
VisPruner (ICCV'25) & 43.9 & 9.5 & 13.9 & 33.9 \\
DTC (CVPR'25)       & 39.9 & 9.0 & 13.4 & 33.7 \\
\textbf{3DZip (Ours)}        & \textbf{49.2} & \textbf{9.6} & \textbf{14.3} & \textbf{34.4} \\
\midrule

\rowcolor{orange!8}
\multicolumn{5}{c}{\emph{Retain 64 Tokens ↓(4.5\%)}} \\
VisPruner (ICCV'25) & 38.9 & 9.0 & 13.6 & 33.2 \\
DTC (CVPR'25)       & 36.0 & 8.8 & 13.3 & 33.1 \\
\textbf{3DZip (Ours)}      & \textbf{45.7} & \textbf{9.3} & \textbf{14.0} & \textbf{33.8} \\
\bottomrule
\end{tabular}
}
\label{tab:3dcap} 
\end{table}

\begin{table}[t!]
\caption{\textbf{Additional captioning-oriented metrics on ScanQA.}
We report BLEU-4, METEOR, ROUGE-L, and CIDEr scores under different token budgets. 3DZip consistently outperforms existing token compression methods across all metrics.}
\centering
\renewcommand{\arraystretch}{0.9}
\resizebox{0.7\columnwidth}{!}{%
\begin{tabular}{lccccc}
\toprule
\textbf{Method} & \textbf{  CIDEr      } & \textbf{ BLEU-4 } & \textbf{  METEOR  } & \textbf{  ROUGE-L  } & \textbf{EM} \\
\midrule
\rowcolor{orange!8}
\multicolumn{6}{c}{\emph{All 1410 Tokens (Baseline)}} \\
LLaVA-3D (ICCV'25) & 85.5 & 11.2 & 17.1 & 43.6 & 26.5 \\
\midrule

\rowcolor{orange!8}
\multicolumn{6}{c}{\emph{Retain 128 Tokens ↓(9.1\%)}} \\
FastV(ECCV'24) & 71.8 & 8.9 & 14.6 & 37.4 & 21.9 \\
SparseVLM(ICML'25) & 69.2 & 9.7 & 14.0 & 36.3 & 21.8 \\
VisionZip(CVPR'25) & 73.1 & 10.9 & 14.7 & 37.7 & 22.2 \\
VisPruner (ICCV'25) & 73.1 & 10.3 & 14.8 & 37.7 & 22.3 \\
Voxelization        & 75.8 & 10.5 & 15.3 & 39.0 & 23.6 \\
DTC (CVPR'25)       & 72.6 & 10.4 & 14.8 & 37.8 & 22.1 \\
\textbf{3DZip (Ours)}       & \textbf{78.2} & \textbf{11.4} & \textbf{15.6} & \textbf{40.3} & \textbf{24.2} \\
\midrule

\rowcolor{orange!8}
\multicolumn{6}{c}{\emph{Retain 64 Tokens ↓(4.5\%)}} \\
FastV(ECCV'24) & 68.9 & 8.0 & 14.0 & 36.0 & 21.1 \\
SparseVLM(ICML'25) & 67.4 & 8.5 & 13.8 & 35.5 & 20.9 \\
VisionZip(CVPR'25) & 66.0 & 8.3 & 13.7 & 34.9 & 20.0 \\
VisPruner (ICCV'25) & 71.6 & 9.0  & 14.5 & 37.1 & 21.9 \\
Voxelization        & 70.9 & 10.0 & 14.6 & 37.1 & 21.9 \\
DTC (CVPR'25)       & 69.5 & 8.7  & 14.3 & 36.6 & 21.3 \\
\textbf{3DZip (Ours)}        & \textbf{76.1} & \textbf{10.7} & \textbf{15.3} & \textbf{39.0} & \textbf{23.3} \\
\midrule

\rowcolor{orange!8}
\multicolumn{6}{c}{\emph{Retain 32 Tokens ↓(2.3\%)}} \\
FastV(ECCV'24) & 65.5 & 8.1 & 13.6 & 34.6 & 19.9 \\
SparseVLM(ICML'25) & 65.5 & 7.8 & 13.4 & 34.8 & 20.4 \\
VisionZip(CVPR'25) & 64.9 & 7.7 & 13.7 & 34.7 & 19.6 \\
VisPruner (ICCV'25) & 67.4 & 8.0  & 14.0 & 35.9 & 20.9 \\
Voxelization        & 66.9 & 8.6  & 13.8 & 35.4 & 20.7 \\
DTC (CVPR'25)       & 67.7 & 9.0  & 14.1 & 35.8 & 20.6 \\
\textbf{3DZip (Ours)}        & \textbf{70.5} & \textbf{9.7} & \textbf{14.5} & \textbf{37.0} & \textbf{21.9} \\
\bottomrule
\end{tabular}
}
\vspace{-10pt}
\label{tab:other_metric}
\end{table}

\subsection{Evaluation on 3D Dense Captioning}
\label{sec:3dcap}

In addition to the 3D question answering task, we further evaluate our method on the 3D dense captioning task. 
3D dense captioning aims to generate natural language descriptions for objects in a 3D scene, where object-level region proposals are provided together with their 3D coordinates.
We evaluate our method on the Scan2Cap dataset~\cite{chen2021scan2cap}, which is built on ScanNet and contains human-annotated captions for object instances in indoor scenes. 
The evaluation is conducted on the Scan2Cap validation split consisting of 7,023 samples across 102 scenes.
For a fair comparison, we adopt the same LLaVA-3D model and apply 3DZip at the same token compression stage used in the 3D question answering experiments, without modifying the original model architecture or training procedure. 
As shown in Table~\ref{tab:3dcap}, 3DZip consistently outperforms existing token compression methods across multiple captioning metrics.

\subsection{Additional Metrics}
\label{sec:other_metric}

In addition to exact match (EM), we report additional captioning-oriented metrics on ScanQA, including BLEU-4, METEOR, ROUGE-L, and CIDEr, to provide a more comprehensive evaluation of the generated responses. These metrics capture complementary aspects of generation quality, such as n-gram overlap and semantic similarity, and therefore provide a more detailed assessment of how token compression affects language generation.
As shown in Table~\ref{tab:other_metric}, 3DZip consistently achieves the best performance across all metrics and token budgets. Notably, the improvements remain stable even under aggressive compression (e.g., 64 and 32 tokens), indicating that the proposed method preserves not only answer correctness but also the semantic fidelity of generated responses. 
These results further demonstrate that feature-diversity-driven token selection maintains informative visual tokens that are critical for both reasoning accuracy and natural language generation quality.

\subsection{Additional Qualitative Results}
\label{sec:qualitative}

\begin{figure}
    \centering
    \includegraphics[width=0.95\linewidth]{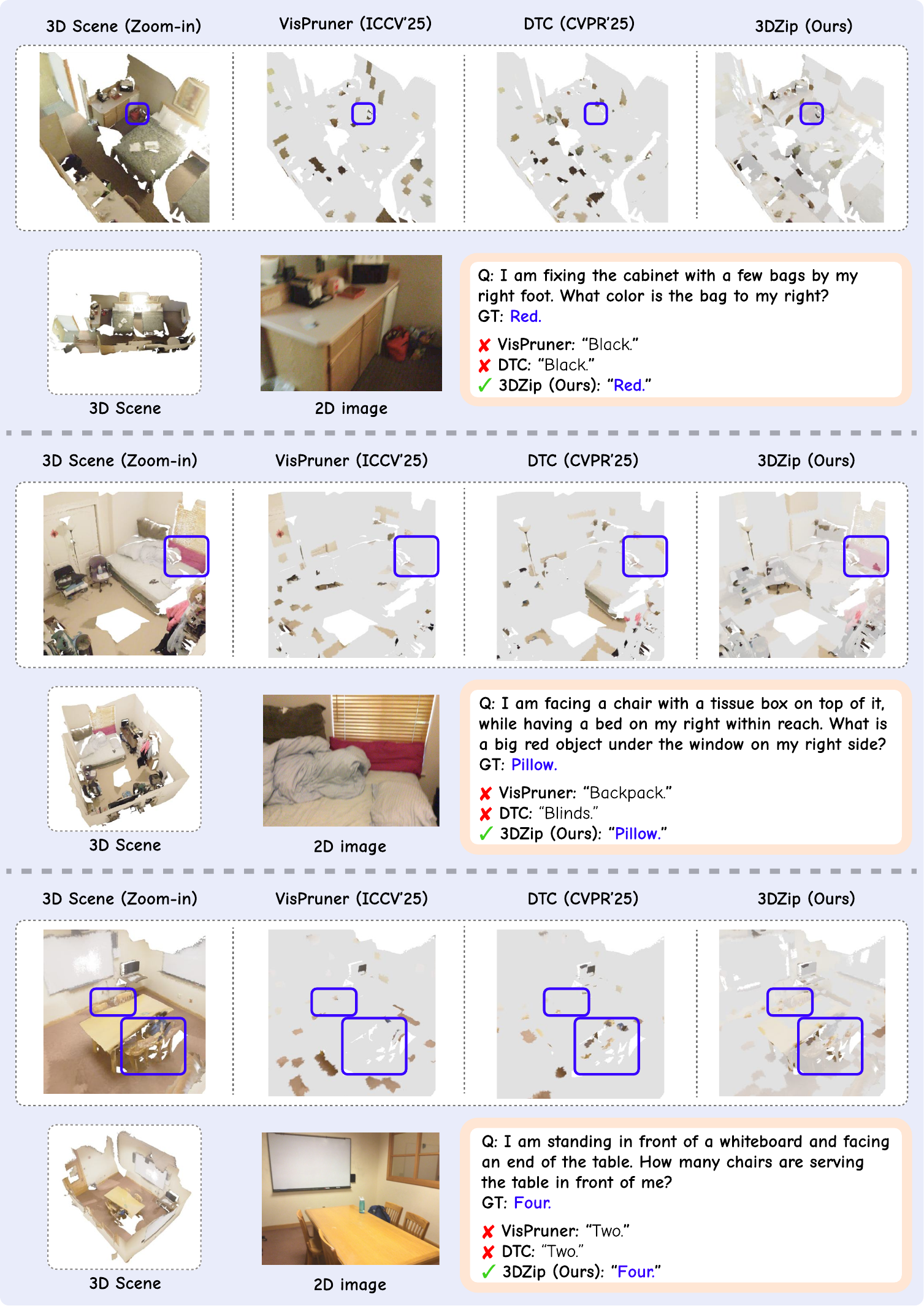}
    \caption{\textbf{Additional qualitative examples on SQA3D.} This figure illustrates the retained 3D tokens produced by 3DZip across different indoor scenes.}
    \label{fig:qualitative_sqa3d}
\end{figure}

\begin{figure}
    \centering
    \includegraphics[width=0.95\linewidth]{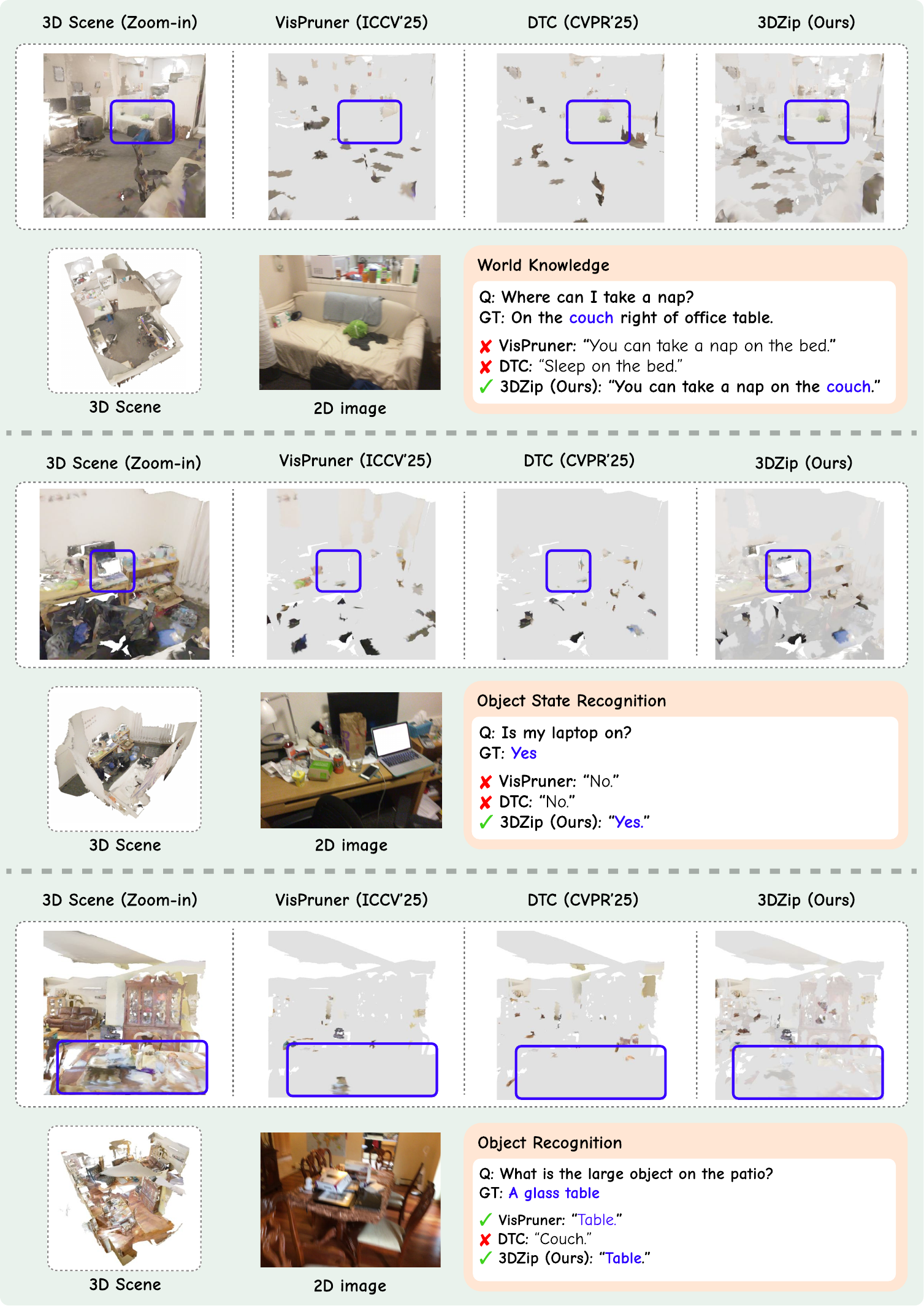}
    \caption{\textbf{Additional qualitative examples on OpenEQA.} This figure illustrates the retained 3D tokens produced by 3DZip across different indoor scenes.}
    \label{fig:qualitative_openeqa}
\end{figure}

\begin{figure}
    \centering
    \includegraphics[width=0.95\linewidth]{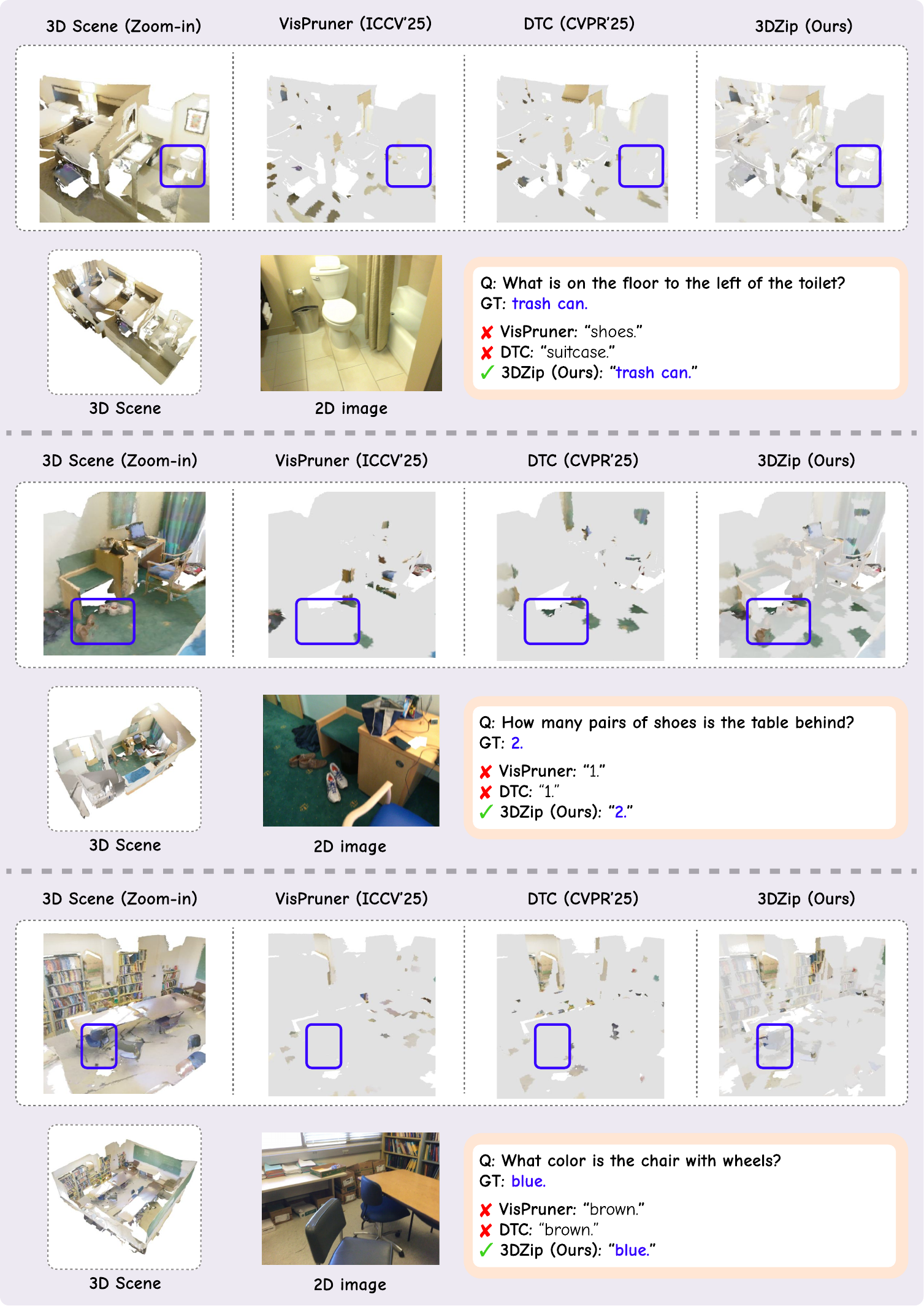}
    \caption{\textbf{Additional qualitative examples on ScanQA.} This figure illustrates the retained 3D tokens produced by 3DZip across different indoor scenes.}
    \label{fig:qualitative_scanqa}
\end{figure}

We provide additional qualitative results of retained 3D tokens in Figs.~\ref{fig:qualitative_sqa3d},~\ref{fig:qualitative_openeqa}, and~\ref{fig:qualitative_scanqa}. 3DZip tends to retain semantically diverse tokens across the scene while incorporating complementary context through spatially-constrained merging.

\newpage



\begin{table}[t]
\centering
\caption{\textbf{Point-level redundancy under voxel aggregation.}
PLR (Eq.~\eqref{eq:plr_delta}) is the mean number of tokens surviving per
physical point; $1.0$ denotes no duplication. PLR drops monotonically with
voxel size $\delta$ while object purity degrades, and EM peaks at
$\delta{=}0.2$\,m.}
\resizebox{0.6\linewidth}{!}{%
\begin{tabular}{ccccc}
\toprule
$\delta$ & PLR $\downarrow$ & Obj. purity $\uparrow$ & Tok./vox. & EM@64 $\uparrow$ \\
\midrule
w/o              & 4.74          & N/A            & N/A   & 51.8 \\
0.10\,m          & 3.09          & \textbf{0.975} & 3.1   & 52.6 \\
\rowcolor{gray!8}
\textbf{0.20\,m} & 2.50          & 0.915          & 9.5   & \textbf{52.8} \\
0.30\,m          & \textbf{2.25} & 0.847          & 19.4  & 52.1 \\
\bottomrule
\end{tabular}}
\label{tab:plr_appendix}
\end{table}

\section{Additional Analysis}
\subsection{Point-Level Token Redundancy Analysis}
\label{sec:point_level_redunduncy}

LLaVA-3D back-projects multi-view 2D features into 3D, so the same surface point
observed from overlapping views can produce near-duplicate tokens. We call this
\emph{point-level redundancy} (PLR), which mainly increases the token count
without adding new semantic information. Stage~1 reduces PLR through voxel
aggregation.

We estimate PLR by detecting SuperPoint~\cite{detone2018superpoint} keypoints,
matching them with LightGlue~\cite{lindenberger2023lightglue}, and forming
\emph{tracks} as connected components of the match graph. To focus on multi-view
redundancy, we exclude singleton tracks from the PLR computation and let
$\mathcal{C}$ denote the remaining set of multi-view tracks, where each track
$\tau$ contains repeated observations of one physical point. PLR is defined as
\begin{equation}
\label{eq:plr}
\mathrm{PLR}=\frac{1}{|\mathcal{C}|}\sum_{\tau\in\mathcal{C}}|\tau| .
\end{equation}
After voxelization with side length $\delta$, the remaining redundancy is
\begin{equation}
\label{eq:plr_delta}
\mathrm{PLR}(\delta)=\frac{1}{|\mathcal{C}|}\sum_{\tau\in\mathcal{C}}
\bigl|\{\,\mathrm{vox}_{\delta}(p):p\in\tau\,\}\bigr| .
\end{equation}
This measures how many distinct voxels each repeatedly observed physical point
still occupies on average and is independent of the final token budget. We also
report object purity and tokens/voxel.

Table~\ref{tab:plr_appendix} reports the SQA3D results, with EM evaluated under
a 64-token budget. PLR decreases from $4.74$ to $3.09/2.50/2.25$ for
$\delta{=}0.1/0.2/0.3$\,m, confirming that Stage~1 effectively reduces geometric
duplication. However, object purity drops from $0.975$ to $0.847$, showing the
trade-off between aggregation and object mixing. EM peaks at $\delta{=}0.2$\,m
($52.8$), where voxelization reduces the PLR value by about $47\%$ while
maintaining high purity ($0.915$). The remaining PLR of $2.5$ indicates that
substantial redundancy still remains after voxelization, motivating Stage~2,
which further reduces residual redundancy through DPP-based feature-diverse
anchor selection.

\begin{figure}[t]
    \centering
    \includegraphics[width=1\linewidth]{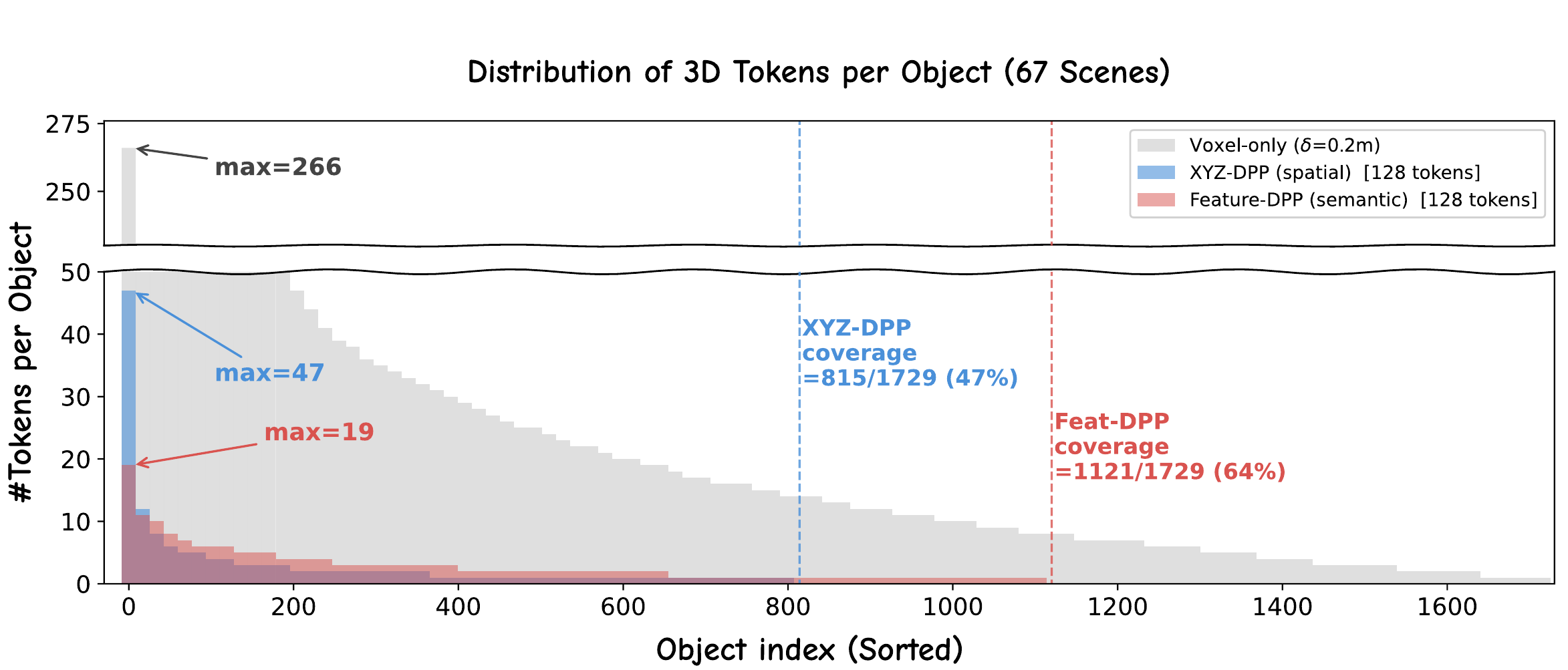}
\caption{
\textbf{Object-level token allocation across selection strategies on the full SQA3D test scenes.}
Per-object token counts for 1,729 foreground object instances across 67 ScanNet scenes from the SQA3D test split. 
The voxel-only distribution (gray) reveals a pronounced long-tail pattern, where a small subset of objects receives a disproportionately large number of tokens.
Spatial sampling (XYZ-DPP) reduces the extreme concentration but still allocates tokens to a limited subset of objects, resulting in 47\% object coverage. 
In contrast, Feature-DPP promotes diversity in feature space and distributes tokens more evenly across object instances, improving object coverage to 64\%.}
    \label{fig:objectcoveragegraph}
\end{figure}

\begin{table*}[t]
    \centering
    \caption{\textbf{Component-wise ablation on SQA3D.}
    (a) Comparison of diversity-promoting strategies for anchor selection.
    (b) Effect of different feature-space distance metrics.
    (c) Effect of the spatial constraint threshold $\tau_g$ in token merging.}
    \resizebox{0.8\columnwidth}{!}{%
    
    \begin{subtable}[t]{0.32\textwidth}
        \centering
        \begin{tabular*}{\linewidth}{@{\extracolsep{\fill}}lc}
        \toprule
        \textbf{Method} & \textbf{EM} \\
        \midrule
        \rowcolor{gray!8}
        \multicolumn{2}{c}{\emph{All 1410 Tokens}} \\
        LLaVA-3D & 55.7 \\
        \midrule
        \rowcolor{gray!8}
        \multicolumn{2}{c}{\emph{Retain 64 Tokens}} \\
        Voxelization &49.8 \\
        DTC (CVPR'25) & 50.2 \\
        \cmidrule(lr){1-2}
        FPS-based & 52.2 \\
        MMDP-based & 52.5 \\
        \textbf{DPP-based} & \textbf{52.8} \\
        \midrule
        \rowcolor{gray!8}
        \multicolumn{2}{c}{\emph{Retain 32 Tokens}} \\
        Voxelization & 47.9\\
        DTC (CVPR'25) & 48.8 \\
        \cmidrule(lr){1-2}
        FPS-based & 51.0 \\
        MMDP-based & 51.0 \\
        \textbf{DPP-based} & \textbf{51.1} \\
        \bottomrule
        \end{tabular*}
        \vspace{2pt}
        \subcaption{}
        \label{tab:diversity_compare}
    \end{subtable}
    \hspace{0.015\textwidth}
    \begin{subtable}[t]{0.32\textwidth}
        \centering
        \begin{tabular*}{\linewidth}{@{\extracolsep{\fill}}lc}
        \toprule
        \textbf{Method} & \textbf{EM} \\
        \midrule
        \rowcolor{gray!8}
        \multicolumn{2}{c}{\emph{All 1410 Tokens}} \\
        LLaVA-3D & 55.7 \\
        \midrule
        \rowcolor{gray!8}
        \multicolumn{2}{c}{\emph{Retain 64 Tokens}} \\
        Voxelization &49.8 \\
        DTC (CVPR'25) & 50.2 \\
        \cmidrule(lr){1-2}
        $\ell_1$ & 52.5 \\
        $\ell_2$ & 52.4 \\
        \textbf{Cosine} & \textbf{52.8} \\
        \midrule
        \rowcolor{gray!8}
        \multicolumn{2}{c}{\emph{Retain 32 Tokens}} \\
        Voxelization & 47.9\\
        DTC (CVPR'25) & 48.8 \\
        \cmidrule(lr){1-2}
        $\ell_1$ & 50.9 \\
        $\ell_2$ & 50.6 \\
        \textbf{Cosine} & \textbf{51.1} \\
        \bottomrule
        \end{tabular*}
        \vspace{2pt}
        \subcaption{}
        \label{tab:ablation_distance}
    \end{subtable}
    \hspace{0.015\textwidth}
    \begin{subtable}[t]{0.32\textwidth}
        \centering
        \renewcommand{\arraystretch}{0.92}
        \begin{tabular*}{\linewidth}{@{\extracolsep{\fill}}lc}
        \toprule
        \textbf{Method} & \textbf{EM} \\
        \midrule
        \rowcolor{gray!8}
        \multicolumn{2}{c}{\emph{All 1410 Tokens}} \\
        LLaVA-3D & 55.7 \\
        \midrule
        \rowcolor{gray!8}
        \multicolumn{2}{c}{\emph{Retain 64 Tokens}} \\
        (w/o Merge) & 52.5 \\
        $\tau_g=1$ & 52.5 \\
        $\tau_g=3$ & 52.7 \\
        \textbf{$\tau_g=5$} & \textbf{52.8} \\
        $\tau_g=7$ & 52.6 \\
        (w/o Spatial-Const.) & 52.3 \\
        \midrule
        \rowcolor{gray!8}
        \multicolumn{2}{c}{\emph{Retain 32 Tokens}} \\
        (w/o Merge) & 50.5 \\
        $\tau_g=1$ & 50.6 \\
        $\tau_g=3$ & 50.9 \\
        \textbf{$\tau_g=5$} & \textbf{51.1} \\
        \textbf{$\tau_g=7$} & \textbf{51.1} \\
        (w/o Spatial-Const.) & 50.8 \\
        \bottomrule
        \end{tabular*}
        \vspace{2pt}
        \subcaption{}
        \label{tab:ablation_constrained}
    \end{subtable}
    }

    \label{tab:main_ablation}
    \vspace{-5pt}
\end{table*}

\subsection{Object-Level Token Allocation Analysis}
\label{sec:object_coverage}
\vspace{2mm}
To further detail the object-level token allocation described in Sec.~1, we analyze object-level token coverage across all 67 ScanNet scenes in the SQA3D test split. Tokens are back-projected into 3D world coordinates and voxelized at $\delta = 0.2$m resolution. To associate tokens with object instances, we use the GT instance annotations provided by ScanNet. ScanNet mesh vertices are discretized at the same resolution and each voxel is assigned the majority-vote instance label of its contained vertices. Background categories (floor, wall, ceiling) are excluded, leaving only foreground objects for analysis.

Fig.~\ref{fig:objectcoveragegraph} plots the per-object token counts for all 1,729 foreground object instances across the 67 ScanNet scenes, sorted in descending order by the voxel-only allocation. The voxel-only distribution confirms the object-level redundancy discussed in Sec.~1: token allocation follows a pronounced long-tail in which the largest object receives up to 266 tokens while most objects receive fewer than 10.
Selecting 128 tokens via XYZ-DPP, which constructs a DPP kernel from 3D voxel coordinates to maximize spatial spread, reduces the maximum number of tokens assigned to a single object from 266 to 47. However, coverage—defined as the fraction of objects receiving at least one token—drops to only 47\% (815/1,729). This occurs because spatial diversity distributes tokens roughly in proportion to an object's geometric extent, causing tokens to concentrate on a limited subset of large objects.

Feature-DPP replaces spatial similarity with cosine similarity over CLIP features and instead promotes diversity in the feature space. As a result, it achieves substantially better coverage at 64\% (1,121/1,729) with a further-reduced maximum of 19 tokens per object by preferentially selecting tokens from semantically distinct objects. 
These results suggest that mitigating object-level redundancy requires considering feature-level diversity rather than relying solely on spatial uniformity. This observation motivates the adaptive token allocation strategy adopted by 3DZip.

\subsection{Comparison of Diversity Strategies}
\label{sec:diversity_comparison}
We adopt DPP as one practical mechanism to promote diversity in feature space. 
To verify that the observed performance gain arises from the diversity objective itself rather than the specific algorithm, 
we compare DPP with two alternative diversity-promoting strategies: 
Farthest Point Sampling (FPS)~\cite{eldar1997fps}, which greedily selects tokens with the largest feature-space distance, 
and the Max-Min Diversity Problem (MMDP)~\cite{resende2010mmdp}, which maximizes the minimum pairwise distance among selected tokens.
As shown in Table~\ref{tab:diversity_compare}, all diversity-based strategies consistently outperform spatial-based baselines, 
indicating that encouraging feature-space diversity is the key factor for effective token allocation in projection-based 3D VLMs. 
Among these approaches, DPP achieves slightly better performance, likely because its determinant-based formulation promotes global diversity among all selected tokens rather than relying on purely greedy pairwise distances.

\begin{table}[t]
\centering
\caption{\textbf{Comparison with 2D Diversity-based method.} 3DZip consistently outperforms DivPrune, demonstrating that spatial awareness is a necessary complement to feature diversity in 3D token compression.}
\setlength{\tabcolsep}{3pt} 
\renewcommand{\arraystretch}{0.5} 
\resizebox{0.7\columnwidth}{!}{%
\begin{tabular}{lccccc}
\toprule
\textbf{Method} & \textbf{3D-Aware} & \textbf{ScanQA} & \textbf{SQA3D} & \textbf{OpenEQA} & \textbf{Rel.} \\
\midrule

\rowcolor{gray!8}
\multicolumn{6}{c}{\emph{All 1410 Tokens}} \\
LLaVA-3D (ICCV'25) & \checkmark & 26.5 & 55.7 & 60.3 & 100.0\% \\
\midrule

\rowcolor{gray!8}
\multicolumn{6}{c}{\emph{Retain 128 Tokens ↓(9.1\%)}} \\
DivPrune (CVPR'25) & \ding{55} & 22.2 & 51.6 & 57.3 & 90.5\% \\
\textbf{3DZip (Ours)} & \checkmark &\textbf{24.2} & \textbf{53.2} & \textbf{58.6} & \textbf{94.7\%} \\
\midrule

\rowcolor{gray!8}
\multicolumn{6}{c}{\emph{Retain 64 Tokens ↓(4.5\%)}} \\
DivPrune (CVPR'25) & \ding{55} & 20.6 & 50.6 & 56.2 & 87.3\% \\
\textbf{3DZip (Ours)} & \checkmark & \textbf{23.3} & \textbf{52.8} &  \textbf{56.7} & \textbf{92.3\%} \\
\midrule

\rowcolor{gray!8}
\multicolumn{6}{c}{\emph{Retain 32 Tokens ↓(2.3\%)}} \\
DivPrune (CVPR'25) & \ding{55} & 19.6 & 49.3 & 54.9 & 84.5\% \\
\textbf{3DZip (Ours)}  & \checkmark & \textbf{21.9} & \textbf{51.1} &  \textbf{55.7} & \textbf{88.9\%} \\
\bottomrule
\end{tabular}%
}
\label{tab:divprune}
\vspace{-10pt}
\end{table}

\vspace{-5pt}
\subsection{Distance Metric Analysis}
\label{sec:distance_metric}
\vspace{-5pt}
As shown in Table~\ref{tab:ablation_distance}, we compare $\ell_1$, $\ell_2$, and cosine similarity to analyze the effect of different distance metrics in feature space during anchor selection. All three metrics show strong performance overall, with cosine similarity achieving the best results. This suggests that measuring similarity based on the direction of normalized feature vectors better captures relationships between tokens.
\subsection{Effect of Spatial Constraints}
\label{sec:spatial_constraints}

We analyze the effect of the spatial constraint threshold $\tau_g$ used in the token merging stage. 
This parameter limits the allowable grid-space distance between anchors and their assigned tokens, preventing spatially distant tokens from being merged into the same anchor.
As shown in Table~\ref{tab:ablation_constrained}, the performance is relatively robust across different values of $\tau_g$. 
A very small threshold restricts contextual aggregation and limits the amount of complementary information that can be merged into anchors, whereas an overly large threshold increases the risk of merging spatially distant tokens that may not belong to the same local region.
In practice, moderate values yield the best results. In our experiments, $\tau_g = 5$ provides the best overall performance across token budgets.

\subsection{Comparison with 2D Diversity-based Token Compression}
\label{sec:divprune}

We compare our method with diversity-based token selection approaches developed for 2D VLMs. DivPrune~\cite{alvar2025divprune} promotes feature diversity during token selection using a Max-Min Diversity Problem (MMDP) objective. However, it considers only feature-space diversity and does not incorporate spatial structure. As shown in Table~\ref{tab:divprune}, 3DZip consistently outperforms DivPrune across all token budgets on ScanQA, SQA3D, and OpenEQA. This result suggests that, for projection-based 3D VLMs, considering spatial structure together with feature diversity leads to more effective token compression.

\begin{table}[t!]
\centering
\caption{\textbf{Robustness of Stage 2 to different Stage-1 aggregation strategies (SQA3D).} Under an identical 64-token budget, adding DPP-based anchor selection (Stage 2) consistently improves EM across all Stage-1 strategies. All variants exclude Stage 3; “+ DPP” denotes adding Stage 2 on top of the corresponding Stage-1 aggregation strategies.}
\renewcommand{\arraystretch}{1.0}
\resizebox{0.4\columnwidth}{!}{%
\begin{tabular}{l@{\hspace{40pt}}c}
\toprule
\textbf{Method}  &\textbf{EM} \\
\midrule
ConceptFusion~\cite{jatavallabhula2023conceptfusion}       & 50.3 \\
\textbf{ConceptFusion + DPP}               & \textbf{52.7}~$_{\color{green!60!black}+2.4}$ \\
\midrule
OpenFusion++~\cite{jin2025openfusion++}         & 49.8 \\
\textbf{OpenFusion++ + DPP}               & \textbf{52.3}~$_{\color{green!60!black}+2.5}$ \\
\midrule
Voxel-mean                               & 49.8 \\
\textbf{Voxel-mean + DPP}                 & \textbf{52.5}~$_{\color{green!60!black}+2.7}$ \\
\bottomrule
\end{tabular}%
}
\label{tab:stage2_robustness}
\end{table}

\subsection{Robustness of Stage 2 to Stage-1 Aggregation Strategies}
\label{sec:stage2_robustness}

We further examine whether feature-diversity-guided anchor selection is tied to the voxel-mean aggregation used in Stage 1. To this end, we replace voxel-mean aggregation with two stronger feature aggregation schemes: ConceptFusion-style feature fusion~\cite{jatavallabhula2023conceptfusion} and OpenFusion++-style area-weighted aggregation~\cite{jin2025openfusion++}. For all variants, we use the same LLaVA-3D backbone, lifted 3D tokens, 64-token budget, and DPP-based anchor selection.

As shown in Table~\ref{tab:stage2_robustness}, adding Stage 2 consistently improves EM across all Stage-1 variants: ConceptFusion from 50.3 to 52.7 (+2.4), OpenFusion++ from 49.8 to 52.3 (+2.5), and voxel-mean aggregation from 49.8 to 52.5 (+2.7). These results show that stronger Stage-1 aggregation does not remove the benefit of feature-diversity selection. While Stage 1 mainly reduces point-level overlap, Stage 2 further mitigates residual object-level redundancy, confirming that the two stages are complementary.

\begin{table}[t]
\centering
\caption{\textbf{Fine-grained performance analysis on ScanQA (EM@64 tokens).} Subsets: \emph{Fine detail} (shape/contour/material/color), \emph{Local struct.} (part-/local structure), \emph{Small obj.} (bottom 10\% by size). \emph{Small obj. purity}: fraction of a voxel's tokens from one small object.}
\setlength{\tabcolsep}{5pt}
\renewcommand{\arraystretch}{1.0}
\resizebox{0.7\columnwidth}{!}{%
\begin{tabular}{lccccc}
\toprule
\multirow{2}{*}{\textbf{Method}}
& \multicolumn{4}{c}{\textbf{EM@64} $\uparrow$}
& \multirow{2}{*}{\makecell{\textbf{Small obj.}\\\textbf{purity} $\uparrow$}} \\
\cmidrule(lr){2-5}
& \textbf{All} & \textbf{Fine.} & \textbf{Local.} & \textbf{Small obj.} & \\
\midrule
\textbf{3DZip (Full)}  & \textbf{23.3} & 30.5 & \textbf{22.4} & 19.5 & 0.426 \\
w/o Voxel (S2+S3)      & 22.8 & \textbf{31.7} & \textbf{22.4} & \textbf{21.6} & N/A \\
Voxel-only (S1)        & 21.9 & 29.9 & 20.1 & 16.9 & 0.070 \\
\bottomrule
\end{tabular}%
}
\label{tab:finegrained}
\end{table}

\subsection{Fine-grained Analysis and Failure Cases}
\label{sec:finegrained}

We analyze fine-grained information loss on three ScanQA subsets: \emph{Fine detail} (shape/contour/material/color), \emph{Local struct.} (part-level or local-neighborhood structure), and \emph{Small obj.} (target objects in the bottom 10\% by size). We additionally report \emph{small-object purity}, defined as the fraction of a voxel's tokens that belong to a single small-object instance.

As shown in Table~\ref{tab:finegrained}, 3DZip achieves the best overall EM (23.3) and preserves \emph{Local struct.} performance (22.4). However, it remains worse than the \emph{w/o Voxel} variant on \emph{Small obj.} and \emph{Fine detail} (19.5/30.5 vs.\ 21.6/31.7), identifying small-object and fine-detail questions as the remaining failure cases. The ablation further indicates that this loss mainly comes from coarse voxelization: \emph{Voxel-only (S1)} performs poorly on these subsets and has low small-object purity (0.070), indicating that different objects are merged into one voxel. Thus, Stage 1 is useful for overall compression but can hurt fine-grained small-object cues.

\begin{table}[t]
\centering
\caption{\textbf{Scene-size dependence on SQA3D.} At a 32-token budget, small scenes favor smaller $\delta$ and tighter $\tau_g$, large scenes the opposite; the fixed setting ($\delta{=}0.2$m, $\tau_g{=}5$) is the best overall trade-off.}
\setlength{\tabcolsep}{6pt}
\renewcommand{\arraystretch}{1.0}
\resizebox{0.45\columnwidth}{!}{%
\begin{tabular}{lccc}
\toprule
\textbf{Param.} & \textbf{Small} & \textbf{Overall} & \textbf{Large} \\
\midrule
\rowcolor{gray!8}
\multicolumn{4}{c}{\emph{Retain 32 Tokens} ($\downarrow$2.3\%)} \\
$\delta$=0.1m & \textbf{54.1} & 50.9 & 43.8 \\
\textbf{$\delta$=0.2m} & 53.6 & \textbf{51.1} & \textbf{44.6} \\
$\delta$=0.3m & 53.5 & 50.6 & 43.8 \\
\midrule
$\tau_g$=1 & \textbf{54.1} & 50.6 & 43.8 \\
$\tau_g$=3 & 53.9 & 50.9 & 44.9 \\
\textbf{$\tau_g$=5} & 53.6 & \textbf{51.1} & 44.6 \\
$\tau_g$=7 & 53.6 & \textbf{51.1} & \textbf{45.9} \\
\bottomrule
\end{tabular}%
}
\label{tab:scene_hyperparam}
\end{table}

\subsection{Effect of Scene Size on Hyperparameters}
\label{sec:scenesize}

We analyze the effect of scene size on the SQA3D dataset. We group scenes by volume and compare the bottom 10\% and top 10\% as small and large scenes, respectively. These two groups differ substantially in density: small scenes are densely packed with 1.30 objects/m$^3$, whereas large scenes are much sparser with 0.37 objects/m$^3$.

As shown in Table~\ref{tab:scene_hyperparam}, small scenes favor a smaller $\delta$ and a tighter $\tau_g$ ($\delta{=}0.1$m, $\tau_g{=}1$ at 32 tokens), while large scenes favor $\delta{=}0.2$m with a looser constraint ($\tau_g{=}7$). This is consistent with the density difference, as a larger $\delta$ or looser $\tau_g$ is more likely to merge tokens from different objects in densely packed small scenes. Overall, the fixed setting ($\delta{=}0.2$m, $\tau_g{=}5$) remains the best trade-off, suggesting scene-adaptive hyperparameters as a promising extension.

\section{Method Details}
\label{sec:method_detail}
\subsection{DPP Algorithm}
\label{sec:dpp_algo}

We provide additional details of the greedy Determinantal Point Process (DPP) anchor selection used in Stage 2 of the main paper. 
DPP is employed as a practical mechanism to encourage feature-space diversity when selecting representative anchor tokens.

Given the normalized voxel features $\{\hat{\mathbf{f}}_k\}_{k=1}^{N_v}$, we construct a cosine similarity kernel $\mathbf{L} \in \mathbb{R}^{N_v \times N_v}$ where $L_{kl} = \hat{\mathbf{f}}_k^\top \hat{\mathbf{f}}_l$. 
Since the features are $\ell_2$-normalized, the resulting kernel is positive semi-definite and captures angular similarity between tokens.

Anchor selection is performed by approximately maximizing $\log \det(L_{\mathcal{A}})$ using the fast greedy algorithm with Cholesky updates~\cite{chen2018fast}. 
Intuitively, this objective favors subsets of tokens that span diverse directions in feature space, thereby suppressing redundant tokens while preserving semantically distinct ones.

Concretely, we maintain a residual score $d_k$ initialized as $d_k = L_{kk}$ and a Cholesky factor $\mathbf{C} \in \mathbb{R}^{K \times N_v}$. 
At iteration $t$, the token with the largest residual score is selected:
\begin{equation}
j^* = \arg\max_{k \notin \mathcal{A}} d_k.
\end{equation}

The Cholesky factor and residual scores are then updated as
\begin{equation}
\mathbf{C}_t =
\frac{\mathbf{L}_{j^*} - \sum_{s=1}^{t-1} C_{s,j^*} \mathbf{C}_s}{\sqrt{d_{j^*}}},
\qquad
d_k \leftarrow d_k - C_{t,k}^2.
\end{equation}

The selected token is excluded from future iterations by setting $d_{j^*}=-\infty$, and residual scores are clamped to zero to ensure numerical stability. 
After $K$ iterations, the resulting set $\mathcal{A}$ forms the anchor tokens used in the subsequent merging stage.

\subsection{DPP Runtime Analysis}
\label{sec:dpp_complexity}

\begin{table}[t]
\centering
\caption{\textbf{Latency comparison of token selection methods at $K=128$.} 
We report the selection algorithm time, total inference latency, and the proportion of the selection overhead relative to the total time.}
\label{tab:dpp_latency}
\resizebox{0.6\columnwidth}{!}
{%
\begin{tabular}{lcccc}
\toprule
\textbf{\shortstack{Method}} &
\textbf{\shortstack{Retain\\Tokens}} &
\textbf{\shortstack{Algorithm\\Latency (ms)}} &
\textbf{\shortstack{Total\\Latency (ms)}} &
\textbf{\shortstack{Overhead\\(\%)}} \\
\midrule
DTC & 128 & 36 & 196 & 18.4 \\
\textbf{DPP (Ours)} & 128 & \textbf{18} & \textbf{178} & \textbf{10.1} \\
\bottomrule
\end{tabular}
}
\vspace{-5pt}
\end{table}

Table~\ref{tab:dpp_latency} reports the runtime of the token selection stage under the same token budget ($K=128$). 
Our DPP-based selection takes 18\,ms, whereas the iterative selection used in DTC takes 36\,ms, resulting in a smaller selection overhead relative to the total inference time.
In practice, the overhead of DPP-based selection remains modest because the preceding voxelization stage significantly reduces the number of tokens before anchor selection.

\vspace{-3pt}
\section{Experimental Details}
\label{sec:sup_exp}

\subsection{Experimental Setup}
\label{sec:sup_exp_details}

\subsubsection{LLaVA-3D}

LLaVA-3D~\cite{zhu2024llava} extends the LLaVA framework to 3D scene understanding by projecting multi-view RGB-D observations into a shared 3D coordinate system. 
It extracts 2D patch features using a CLIP visual encoder and augments them with 3D positional embeddings derived from depth and camera parameters, producing geometry-aware 3D tokens.
LLaVA-3D originally employs two token compression strategies: Voxelization Pooling, which aggregates tokens spatially within voxels, and FPS Pooling, which selects a fixed number of tokens via farthest point sampling. 
In our experiments, we use the LLaVA-v1.5-7B backbone and adopt Voxelization Pooling as the baseline token construction stage. For fair comparison, all 2D token pruning baselines are applied after this voxelization step. Since the resulting token count varies across scenes, we report the per-dataset average token count as the baseline. 
\vspace{-3mm}
\subsubsection{Video-3D-LLM}

Video-3D-LLM~\cite{zheng2025video3d} models 3D scenes as video sequences and injects 3D positional information into per-frame visual features. 
It uses a SigLIP~\cite{tschannen2025siglip} visual encoder and a Qwen2-7B~\cite{yang2024qwen2technicalreport} language backbone.
3D coordinates obtained from back-projected depth maps are converted into sinusoidal encodings and added to the video features, while frames are selected using a greedy Maximum Coverage Sampling strategy.
After the MLP projector, per-frame 2D average pooling reduces spatial resolution before applying 3D positional embeddings. 
Following the same training-free principle as in the main experiments, we apply 3DZip after all per-frame token construction operations, immediately before the tokens are fed into the LLM.
\vspace{-3mm}
\subsubsection{SR-3D}

SR-3D~\cite{sr3d} is a 3D-aware VLM built upon the NVILA-Lite-8B~\cite{liu2025nvila} backbone that supports both single-view and multi-view inputs within a unified architecture. 
It leverages a pretrained 2D foundation model and introduces a canonical positional feature that enables effective transfer from single-view pretraining to multi-view 3D reasoning.
Because the MLP projector includes a $3\times3$ convolutional downsampling layer, token compression must be applied after the projector stage. 
Accordingly, we insert 3DZip immediately after the projector output, ensuring that the compression process remains training-free and does not alter the original model architecture.
\vspace{-3mm}

\subsection{Dataset Details}
\label{sec:dataset_details}

\subsubsection{SQA3D}
We evaluate on the test split of SQA3D~\cite{ma2022sqa3d}, which contains 3{,}519 questions across 67 ScanNet scenes.
SQA3D is a situated question answering benchmark where each question is grounded in a specific agent position and orientation within a 3D scene (e.g., ``I am standing by the ottoman facing a couple of toolboxes. What instrument in front of me is ebony and ivory?'').
Questions span six categories: What (26.6\%), Is/Are (33.1\%), How many (12.3\%), What color (6.0\%), Which (10.0\%), and Other (12.1\%).
Evaluation is based on EM   .

\subsubsection{OpenEQA}
OpenEQA~\cite{majumdar2024openeqa} consists of two subsets: ScanNet (1{,}079 questions, 89 scenes) and HM3D (557 questions, 63 scenes).
The HM3D subset requires official episode trajectory states to render frames from the exact camera viewpoints used in the benchmark.
As these states are no longer publicly available, we evaluate on the ScanNet subset only for fair comparison.

\subsubsection{ScanQA}
We evaluate on the validation split of ScanQA~\cite{azuma2022scanqa}, which contains 4{,}675 questions across 71 ScanNet scenes.
Each question asks about object attributes, spatial relationships, or scene properties, and is paired with multiple reference answers. Following standard practice, we report BLEU-4, METEOR, ROUGE-L, and CIDEr, computed against all reference answers.

\subsection{LLM-based Evaluation Prompt}
\label{sec:llm_eval}
As shown in Fig.~\ref{fig:prompt}, the LLM evaluates the predicted answer by comparing it with the ground-truth answer and additional acceptable answers, and outputs a score between 1 and 5.

\begin{figure}[t]
\centering
\fbox{
\begin{minipage}{0.9\linewidth}
\scriptsize
\setstretch{0.9}
\textbf{OpenEQA Evaluation Prompt}

You are an AI assistant who will help me evaluate the response given the question, the correct answer, and extra answers that are also correct.  
Output a single integer between 1 and 5 (inclusive).

5 = the response perfectly matches the answer or any of the extra answers \\
1 = the response is completely different from the answer and all extra answers

\medskip
\textbf{Example 1}\\
Question: Is it overcast?\\
Answer: no\\
Extra Answers: ['doesn't look like it', 'no', 'it's sunny']\\
Response: yes\\
Your mark: 1

\medskip
\textbf{Example 2}\\
Question: Who is standing at the table?\\
Answer: woman\\
Extra Answers: ['a woman', 'a lady', 'woman']\\
Response: Jessica\\
Your mark: 3

\medskip
\textbf{Example 3}\\
Question: Are there drapes to the right of the bed?\\
Answer: yes\\
Extra Answers: ['yes, there are drapes', 'yeah', 'the drapes are to the right of the king bed']\\
Response: yes\\
Your mark: 5

\medskip
\textbf{Your Turn}\\
Question: \{question\}\\
Answer: \{answer\}\\
Extra Answers: \{extra\_answers\}\\
Response: \{prediction\}

\end{minipage}
}
\vspace{-3mm}
\caption{Prompt used for LLM-based evaluation in OpenEQA.}
\label{fig:prompt}
\end{figure}

\end{document}